\documentclass[review,number]{elsarticle}
\usepackage{hyperref,amsmath,amssymb,amsthm}
\usepackage{subfig,algorithm,algorithmic,footnote}
\usepackage{epstopdf,color}
\journal{} %{Computer Vision and Image Understanding}
\begin{document}
\begin{frontmatter}
\title{COROLA: A Sequential Solution to Moving Object Detection Using Low-rank Approximation}
%\title{A Sequential Solution to Moving Object Detection by Contiguous Outliers Representation Using Online Low-rank Approximation (COROLA)}

\author{Moein Shakeri, Hong Zhang}
\address{Department of Computing Science\\ University of Alberta, Edmonton, Alberta, CANADA\\}
\address{shakeri,hzhang@ualberta.ca}

\begin{abstract}
Extracting moving objects from a video sequence and estimating the background of each individual image are fundamental issues in many practical applications such as visual surveillance, intelligent vehicle navigation, and traffic monitoring. Recently, some methods have been proposed to detect moving objects in a video via low-rank approximation and sparse outliers where the background is modeled with the computed low-rank component of the video and the foreground objects are detected as the sparse outliers in the low-rank approximation. Many of these existing methods work in a batch manner, preventing them from being applied in real time and long duration tasks. To address this issue, some online methods have been proposed; however, existing online methods fail to provide satisfactory results under challenging conditions such as dynamic background scene and noisy environments. In this paper, we present an online sequential framework, namely contiguous outliers representation via online low-rank approximation (COROLA), to detect moving objects and learn the background model at the same time. We also show that our model can detect moving objects with a moving camera. Our experimental evaluation uses simulated data and real public datasets to demonstrate the superior performance of COROLA to the existing batch and online methods in terms of both accuracy and efficiency.

\end{abstract}

\end{frontmatter}

\section{Introduction}
\label{section1}
Moving object detection and background estimation are fundamental in various applications of computer vision and robotics such as visual surveillance~\cite{int_app1}, traffic monitoring~\cite{int_app2}, vehicle tracking and navigation~\cite{int_app3}, and avian protection~\cite{int_app4}. Many methods have been proposed to extract objects from a sequence of images with a stationary camera~\cite{int1},~\cite{int2} or with a moving camera~\cite{int3},~\cite{int4},~\cite{int5}. These methods can be grouped into several categories. Motion-based methods~\cite{int_motion},~\cite{int_motion2} use motion information of the image pixels to separate the foreground from the background. These methods work based on the assumption that foreground objects move differently from the background. Therefore it is possible for these methods to classify pixels according to their movement characteristics even in the case of significant camera motion. However, these methods require point tracking to identify the foreground, which can be difficult especially with large camera motion~\cite{int_segmentation}. In addition, they are limited in terms of dealing with dynamic background or noisy data~\cite{int_segmentation2}.

Another popular category for moving object detection methods is background subtraction~\cite{int6}, which compares the pixels of an image with a background model and considers those that differ from the background model as moving objects. Thus, building a background model plays a critical role in background subtraction methods. Conventional algorithms for background modelling include single Gaussian distribution~\cite{int_SGM}, Gaussian mixture model~\cite{int_MOG}, and kernel density estimation~\cite{int_KDE}. These methods model the background for each pixel independently and so they are not robust against global variations such as illumination changes.

Recently a new approach to background modelling, namely low-rank matrix approximation, has been developed~\cite{int_lowrank,Related_RPCA}. Methods in this approach follow the basic idea from~\cite{Eigenbackground}. Oliver {\it{et al.}}~\cite{Eigenbackground} proposed Eigenbackground subtraction using PCA~\cite{PCA} (principal component analysis) to model the background and detect moving objects. It is based on the observation that the underlying background images should be unchanged and the composed matrix of vectorized background images can be naturally modeled as a low-rank matrix. Extending this idea, current methods exploit the fact that the background model in an image sequence can be defined by those pixels that are temporally linearly correlated~\cite{int7}. By capturing the correlation between images one can naturally handle global variations. Algebraically speaking, if an image is vectorized in a column and all images are concatenated into a 2D matrix, then the columns are dependent and its low-rank approximation matrix represents the background model of the images. As a result, the background modeling problem is converted to the low-rank approximation problem. In general, by decomposing an input matrix of vectorized images into a low-rank matrix and a sparse matrix, the low-rank and sparse matrices correspond to the background model and the foreground objects in the image sequence respectively. Our COROLA algorithm described in this paper adopts the low-rank approximation approach. We will detail representative algorithms in this approach in Section~\ref{section2}.

Most of the existing background subtraction algorithms based on low-rank approximation operate in a batch manner; i.e., all images whose background model is to constructed are first collected and then used to build a data matrix whose low-rank approximation is computed.  This unfortunately limits the application of the low-rank approximation approach in terms of its efficiency and accuracy. Although existing online methods via low-rank approximation have addressed the efficiency issue to some extent, they are not robust against dynamic and noisy background. In this paper, we offer an algorithm, COROLA,  that performs  low-rank approximation in a sequential manner so that its computational complexity does not grow with the number of images in the sequence.  In addition, through image registration, our algorithm is able to handle the case of a moving camera due to the adaptive nature of the background model that is being learned.
%Specifically, this paper deals with the problem of online moving object detection and our method provides an online formulation to compute the background model and the foreground mask for input images at the same time for both stationary and moving cameras.
The main contributions of this paper are as follows.

1. We propose an online formulation of the low-rank approximation algorithm for foreground object detection. The proposed formulation enables online application without requiring an entire image sequence, as in the batch formulation and is more robust than existing online methods for dynamic background scene or noisy environment.

2. COROLA uses a fixed window of images to perform low-rank approximation and so it is appropriate for continuous operation, which cannot be achieved by the batch formulation due to matrix decomposition and memory storage.

3. In the case of significant camera motion, a batch formulation has the limitation that the first and the last images of a sequence must be similar to find the low-rank matrix. However, in the case of a moving camera, there is in general no similarity between the first and the last images in a sequence. Our proposed COROLA algorithm does not require a stationary background.

The remainder of the paper is organized as follows. Related works on foreground detection via low-rank and sparse decomposition are summarized in Section~\ref{section2}. Section~\ref{section3} explains the details of COROLA for foreground detection and background estimation, followed by the introduction of our online formulation via greedy bilateral sketch~\cite{bilateral_sketch}. Experimental results and discussion are presented in Section~\ref{section4}, and concluding remarks in Section~\ref{section5}.

\section{Foreground Detection via Low Rank and Sparse Decomposition}
\label{section2}
In recent years, many algorithms have been developed for foreground detection based on low-rank matrix approximation with robust principal component analysis (RPCA)~\cite{Related_RPCA}. RPCA decomposes a given matrix $D$ into low-rank matrix $L$ and sparse matrix $S$ called outliers. Different techniques exist for low-rank approximation including principal component pursuit (PCP)~\cite{int7}, augmented Lagrangian multiplier (ALM)~\cite{Related_ALM}, linearized alternating direction method with an adaptive penalty (LADMAP)~\cite{Related_LADMAP}, and singular value thresholding (SVT)~\cite{Related_SVT}. All of these techniques need all the data in order to perform batch optimization that computes the low-rank matrix and the sparse outliers. Due to batch processing, the following two problems occur: {\it{memory storage}} and {\it{time complexity}}. In continuous monitoring tasks or video processing, if matrix $D$ is built with a large number of images memory storage will be a problem~\cite{Related_memory}. In addition, by increasing the size of the input matrix $D$, time complexity for the matrix decomposition is also increasing.

To address the problem of time complexity, some efficient algorithms have been proposed~\cite{bilateral_sketch,Related_FPCP,GoDec}. Rodrigues and Wohlberg proposed a fast PCP~\cite{Related_FPCP} algorithm to reduce the computation time of SVD in inexact ALM (IALM). The ``Go Decomposition" (GoDec) method, proposed by Zhou {\it{et al.}} computes RPCA using bilateral random projections (BRP)~\cite{GoDec}. Semi-Soft GoDec (SSGoDec) and Greedy SSGoDec methods~\cite{bilateral_sketch} are extensions of GoDec to speedup it. Although these algorithms reduce the computation time of low-rank approximation, they still are not satisfactory for applications such as visual surveillance and robot navigation due to their batch formulation. In many applications, online processing is critical and batch methods are infeasible. One of the best known batch processing algorithms is the ``detecting contiguous outliers in the low-rank representation" (DECOLOR) method~\cite{Related_decolor}. This method uses a priori knowledge of the foreground objects that they should be connected components of relatively small size. Using this constraint in the method, DECOLOR provides promising results; however, due to batch processing, it still suffers from memory storage and time complexity problems.
Furthermore, in the case of a moving camera, the current image is no longer similar to the first images in matrix $D$, and therefore DECOLOR is not able to detect foreground appropriately. In general, batch processing methods cannot operate on a continuous basis and cannot deal with a moving camera. Although DECOLOR has introduced an implementation for moving camera, it only works for short video sequences with small camera motion.

To overcome the limitations of batch processing methods, incremental and online robust PCA methods have developed. He {\it{et al.}}~\cite{GRASTA1} proposed Grassmannian robust adaptive subspace tracking algorithm (GRASTA),which is an incremental gradient descent algorithm on Grassmannian manifold for solving the robust PCA problem. This method incorporates the augmented Lagrangian of $l_{1}$-norm loss function into the Grassmannian optimization framework to alleviate the corruption by outliers in the subspace update at each gradient step. Following the idea of GRASTA, He {\it{et al.}}~\cite{t-GRASTA} proposed transformed GRASTA (t-GRASTA), which iteratively performs incremental gradient descent constrained to the Grassmann manifold in order to simultaneously decompose a sequence of images into three parts: a low-rank subspace, foreground objects, and a transformation such as rotation or translation of the image. This method can be regarded as an extension of GRASTA and RASL~\cite{RASL} (Robust Alignment by Sparse and Low-Rank decomposition) by computing the transformation and solving the decomposition with incremental gradient optimization framework. To improve the accuracy of online subspace updates especially for dynamic backgrounds, Xu {\it{et al.}}~\cite{GOSUS} developed an online Grassmannian subspace update algorithm with structured-sparsity (GOSUS) via an alternating direction method of multipliers (ADMM).

To deal with noisy conditions and dynamic background scene, Wang {\it{et al.}}~\cite{PRMF} proposed a probabilistic approach to robust matrix factorization (PRMF) and its online extension for sequential data to obtain improved scalability. This model is based on the empirical Bayes approach and can estimate better background model than GRASTA.
%Guan {\it{et al.}}~\cite{NMF} proposed another incremental approach called online non-negative matrix factorization (OR-NMF) which receives one sample per step and updates the bases via robust stochastic approximation.
Recently, Feng {\it{et al.}}~\cite{Related_OR-PCA} proposed an online robust principal component analysis via stochastic optimization (OR-PCA). This method does not need to remember all the past samples and uses one sample at a time by a stochastic optimization. OR-PCA reformulates a nuclear norm objective function by decomposing to an explicit product of two low-rank matrices, which can be solved by a stochastic optimization algorithm. Javed {\it{et al.}}~\cite{Related_ORPCA_MRF} used this technique for online foreground detection. Their method first extracts outliers from each image using OR-PCA and then uses Markov Random Field (MRF) to improve the quality of foreground segmentation. However, they did not solve the problem of foreground detection within a unified single optimization framework, i.e., MRF is only applied once to improve the outliers of OR-PCA and without alternating learning to update the OR-PCA. As a result, the reported performance is not competitive with respect to those in the literature.

\subsection{Relation of our method to other methods}
\label{section2_1}
Since our COROLA method uses the sparsity and connectedness terms of DECOLOR method and estimates the background model using sequential low-rank approximation with the help of OR-PCA, we present a summary of these two methods and in the next Section we describe our COROLA method that extends the two methods.
\subsubsection{DECOLOR}
DECOLOR is a formulation that integrates the outlier support and the estimated low-rank matrix in a single optimization problem, for joint object detection and background learning. Specifically, it works by solving the following minimization:
\begin{equation}
\label{eq:Decolor}
\begin{split}
\min_{L,S}\frac{1}{2}\|\mathcal P_{S^{\perp}}(D-L)\|_{F}^{2}+\beta_{2}\|S\|_{1}+\gamma\|\Phi(S)\|_{1}\\
 s.t.\,\,\, rank(L)\leq r, \,\,\,\,\,\,\,\,\,\,\,\,\,\,\,\,
\end{split}
\end{equation}
where $D,L, and S$ are the matrix of vectorized images, estimated background images, and outlier support, respectively. $S$ in~(\ref{eq:Decolor}) is binary and its elements are 1 for outliers. $S^{\perp}$ is the complement of $S$ and its elements are 1 for background pixels of the images. $\Phi(S)$ means the difference between neighboring pixels and therefore the last term of the above minimization encourages connectedness of outliers. Zhou {\it{et al.}}~\cite{Related_decolor} solved the first term of~(\ref{eq:Decolor}) with its constraint using an alternating algorithm (SOFT-IMPUTE)~\cite{soft_impute}. They then solved the rest of the minimization problem by Markov Random Field (MRF)~\cite{MRF}. This two-step optimization is iterated until convergence. Although this method provides promissing results, it still suffers from memory storage and time complexity problems in large datasets and, due to batch processing, it is not appropriate to operate on a continuous basis. Furthermore, in the case of a moving camera, DECOLOR only works for short video sequences with small camera motion and cannot deal with a moving camera in general.
\subsubsection{OR-PCA}
OR-PCA solves stochastic optimization sequentially, processing one sample at a time and producing a solution that is equivalent that of the batch RPCA. As a result, its computation cost is independent of the number of samples. OR-PCA solves the following minimization problem:
\begin{equation}
\label{eq:OR-PCA}
\begin{split}
\min_{U,V}\frac{1}{2}\|(D-UV-E)\|_{F}^{2}+\frac{\lambda_{1}}{2}(\|U\|_{F}^{2}+\|V\|_{F}^{2})+\lambda_{2}\|E\|_{1}\\
\end{split}
\end{equation}
where $U$,$R$, and $E$ are the basis, coefficient, and sparse error matrices. Feng {\it{et al.}}~\cite{Related_OR-PCA} solved~(\ref{eq:OR-PCA}) in an online manner for one sample per time by two iterative updating parts. First, the coefficients and the sparse error for each new sample is updated by the previous basis. Then, the basis is updated using the new sample, updated coefficients, and sparse errors.

In this paper, extending the work of DECOLOR and OR-PCA, we introduce a novel non-convex closed-form formulation for detection of moving objects named (COROLA). It solves the challenges of memory storage and time complexity of~\cite{Related_decolor} and provides more accurate results than~\cite{Related_OR-PCA}, especially in noisy environments. COROLA is also able to extract moving objects using a moving camera on a continuous basis, which cannot be achieved in general by a batch processing method especially in the case of large camera motion.

\section{Online Moving Object Detection by COROLA}
\label{section3}
In this section, we focus on online detection of moving objects for both static and moving cameras. We first formulate the problem of background modelling and foreground object detection and then describe in detail our COROLA algorithm, which computes the low-rank approximation and foreground detection sequentially.

\subsection{Notations and Formulation}
Let $X\in R^{m}$ be a vectorized image and $X_{j}$ be the $j^{th}$ image in a sequence, expressed as a column vector of $m$ pixels. Then, $D=[X_{1},...,X_{n}]\in R^{m\times n}$ is a matrix of $n$ images and the $i^{th}$ pixel in the $j^{th}$ image is denoted as $x_{ij}$.
%$X \in R^{m \times r}$ is a matrix of last $r$ observed images, and $r$ is the rank of matrix.
To indicate foreground for an observed image $j$, we use a binary indicator vector $\textsc{s} =[s_1,s_2,...,s_m]^T$ as the foreground support where
\begin{equation}
\label{eq:1}
s_{i} =
  \begin{cases}
   0 & \text{if } i \text{ is background} \\
   1 & \text{if } i \text{ is foreground}
  \end{cases}
\end{equation}
and matrix $S=[\textsc{s}_{1},\textsc{s}_{2},...,\textsc{s}_{n}]$ shows a binary matrix of all images in $D$. Also, we use the function $\mathcal P_{S}(X)\in R^{|\textsc{s}|_{0}}$ to construct a vector of at most $m$ foreground pixels of image $X$. Note that $l_0$-norm $|\textsc{s}|_0$ is the cardinality of $\textsc{s}$ or the number of non-zero elements in $\textsc{s}$.
%\begin{equation}
%\label{eq:2}
%\mathcal P_{S}(X)=\{X_{i}|S_{i}=1\},\,\, i=\{1,...,m\},
%\end{equation}
In a matrix with more than one column, $\mathcal P_{S,:}$ constructs multiple columns each by applying $\mathcal P_{S}$ to a column in the input matrix.
%Also, we use $^{t}X_{j}$ to represent $X_{j}$ in $t^{th}$ iteration.
%$k$ shows the index of last $r$ images in $D$ (last $r$ observed images).
%Based on the above notations, we propose an online formulation as the following equation and we are going to solve it online to be useful for real-time applications.
Now, let $L=UV$. The objective function in~(\ref{eq:Decolor}) can be rewritten as follows.
\begin{equation}
\label{eq:Decolor1}
\begin{split}
\min_{U,V,S}\frac{1}{2}\|\mathcal P_{S^{\perp}}(D-UV)\|_{F}^{2}+\beta_{2}\|S\|_{1}+\gamma\|\Phi(S)\|_{1}\\
 s.t.\,\,\, rank(U)=rank(V)\leq r, \,\,\,\,\,\,\,\,\,\,\,\,\,\,\,\,
\end{split}
\end{equation}
%and by relaxing the constraints based on~\cite{Related_OR-PCA}, the optimization is changed to:
%\begin{equation}
%\label{eq:Decolor2}
%\begin{split}
%\min_{U,V,S}\frac{1}{2}\|\mathcal P_{S^{\perp}}(D-UV)\|_{F}^{2}+\beta_{1}(\|U\|_{F}^2+\|V\|_{F}^2)+\beta_{2}\|S\|_{1}+\gamma\|\Phi(S)\|_{1}\\
%\end{split}
%\end{equation}

With the above notations and equations, and by relaxing the constraints of~(\ref{eq:Decolor1}) based on~\cite{Related_OR-PCA}, the problem of background modelling and foreground object detection via sequential low-rank approximation and contiguous outlier representation solves the following optimization problem for each observed image.
\begin{equation}
\label{eq:3}
\begin{split}
\min_{U,\textsc{v},\textsc{s}}\frac{1}{2}\|\mathcal P_{S}(X-U\textsc{v})\|_{F}^{2}+\beta_{1}\|\mathcal P_{S,:}(U)\|_{F}^2+\beta_{1}\|\textsc{v}\|_{F}^2+\beta_{2}\|\textsc{s}\|_{1}+\gamma\|\Phi(\textsc{s})\|_{1}\\
%s.t.\,\,\, rank(U)\leq r,\,\,\,\,\,\,\,\,\,\,\,\,\,\,\,\,\,\,\,\,\,\,
\end{split}
\end{equation}
%\begin{equation}
%\label{eq:3}
%\begin{split}
%\min_{U,V,S}\frac{1}{2}\|\mathcal P_{S}(X-UV)\|_{F}^{2}+\beta_{2}\|S\|_{1}+\gamma\|\Phi(S)\|_{1}\\
% s.t.\,\,\, rank(U)\leq r,\,\, \|V\|_{0}\leq r \,\,\,\,\,\,\,\,\,\,\,
%\end{split}
%\end{equation}
where $X\in R^{m}$ is an observed image, $r$ is the upper bound on the rank of the basis matrix $U \in R^{m \times r}$, and $\textsc{v} \in R^{r}$ is a coefficient vector. $\Phi(\textsc{s})$ means the difference between neighboring pixels and it is computed by $\|\Phi(\textsc{s})\|_{1}=\underset{(i,k)\in \mathcal E}\sum|s_i-s_k|$ and $\mathcal E$ is the neighborhood clique. Note that the objective function defined in~(\ref{eq:3}) is non-convex and involves both continuous and discrete variables. Since~(\ref{eq:3}) is our online formulation for each input image, the loss over all data would be the cumulative for each image. The first three terms try to compute the low-rank representation of input image $X$ by first expressing it as a linear combination of the background basis $U$ and its coefficient vector $\textsc{v}$, and then penalizing only the foreground pixels using extraction function $\mathcal P_S$. The last two terms of~(\ref{eq:3}) find continuous and small outliers to represent the foreground mask. Specifically, the fourth term imposes a sparsity constraint on the foreground mask $\textsc{s}$; i.e., the foreground pixels should be low in number. The last term imposes a connectivity constraint on mask $\textsc{s}$ to account for correlation between neighboring pixels of an image.
By minimizing~(\ref{eq:3}) we can estimate the best low-rank representation of an input image and detect foreground objects, concurrently. However, solving this joint optimization in one step is difficult. Therefore, people use a two-step alternating optimization procedure by separating it to a low-rank approximation step involving $U$ and $\textsc{v}$, and then a contiguous sparse optimization step involving $\textsc{s}$ to obtain background estimation and foreground detection, performed alternatively. In the first step people treat~(\ref{eq:3}) as minimization over $U$ and $\textsc{v}$, for which we introduce an online approach via the greedy semi-soft GoDec (Gre-SSGoDec) and OR-PCA methods rather than the SOFT-IMPUTE algorithm~\cite{soft_impute} in batch methods. In the second step, minimization over $\textsc{s}$ is conducted. In addition, we use the combination of Gaussian Mixture Model (GMM) and first order MRF with binary labels in the second step to improve the foreground detection performance.
\subsection{Online Low-Rank Approximation}
\label{subsection3_2}
For solving the first step of~(\ref{eq:3}), we describe in this section our sequential method to compute the low rank background model of an image sequence and the foreground as its sparse outliers, in a way that is suitable for continuous and real time operation.
%Let $L=UV$ be a low rank structure of the data matrix $D$ defined in the previous section.  In a batch method, matrix $D$ is factorized to compute $L$. Factorization is computationally expensive for a large $D$, especially for continuous estimation when the number of images in $D$ increases.
%\begin{equation}
%\label{eq:5}
%\begin{split}
%U=DV_{prev}^{T}(V_{prev}V_{prev}^{T})^{\dagger}\\
%V=(U^{T}U)^{\dagger}{U}^{T}D \,\,\,\,\,
%\end{split}
%\end{equation}
%where $V_{prev}$ denotes the variable in the previous iteration and $(.)^{\dagger}$ is the Moore Penrose pseudo inverse. According to (\ref{eq:4}), the column space of $U$ can be represented by arbitrary orthonormal basis for the columns of $DV_{prev}^{T}$and we can compute it as $Q$ via QR-decomposition, where QR($DV_{prev}^{T}$)=$QR$, and thus $U=Q$ and $V=Q^{T}D$~\cite{bilateral_sketch}. Upon convergence, matrix $L_{j=1:n}=UV$ shows the background models, where $L_{j}$ is the low rank of image $X_{j}$. Based on this method we have to factorize matrix $D$, but it is computationally expensive, especially for continuous estimation when the number of images in $D$ increases.
In our sequential formulation, we adopt an online updating approach for optimization over $U$ and $\textsc{v}$. Therefore~(\ref{eq:3}) can be rewritten as:
\begin{equation}
\label{eq:6_0}
\begin{split}
\underset{U,\textsc{v}} {\mathrm{min}}~\frac{1}{2}\|\mathcal P_{S}(X-U\textsc{v})\|_{F}^{2}+ \beta_{1}\|\mathcal P_{S,:}(U)\|_{F}^2 + \beta_{1}\|\textsc{v}\|_{2}^{2}\,\,\,\,\,\\
%s.t.\,\,\, rank(U)\leq r,\,\,
\end{split}
\end{equation}
%as the following steps to be applicable in online and real-time applications:
Since~(\ref{eq:6_0}) updates subspace of $U$ based on foreground mask $\textsc{s}$, we rewrite the objective function for the rest of this section as follows.
\begin{equation}
\label{eq:6_00}
\begin{split}
\underset{\hat{U},\textsc{v}} {\mathrm{min}}~\frac{1}{2}\|\hat{X}-\hat{U}\textsc{v}\|_{F}^{2}+ \beta_{1}\|\hat{U}\|_{F}^2 + \beta_{1}\|\textsc{v}\|_{2}^{2}\,\,\,\,\,\\
%s.t.\,\,\, rank(U)\leq r,\,\,
\end{split}
\end{equation}
where $\hat{U}=\mathcal P_{S,:}(U)$ and $\hat{X}=\mathcal P_{S}(X)$.

{\it{Initialization Step}}: With a small number of images at the beginning of a sequence no fewer than the rank of the background model, we initialize $U$ with a batch method. This enables us to estimate the rank $r$ roughly for the images in the rest of the sequence. Since this step is performed only once, the complexity of using a batch formulation is not an issue.
After the initialization of $U$, for each input sample $X$, we use an incremental approach to solve~(\ref{eq:6_00}) by the following two parts, repeatedly.
%on the remainder of the image sequence or indefinitely on a video feed.
These two parts update $\textsc{v}$, and then $U$ (by updating the subspace of $\hat{U}$) for each sample to build the background model incrementally as follows:

{\it{Part~1}}: Because every two consecutive images in a sequence are similar, we can update coefficient vector $\textsc{v}$ (or $U$) for the current image via background model $U$ (or $\textsc{v}$) computed for the previous image. To update $\textsc{v}$ with the fixed $U$, (\ref{eq:6_0}) becomes:

\begin{equation}
\label{eq:6}
\hat{\textsc{v}}=\underset{\textsc{v}} {\mathrm{argmin}}~\frac{1}{2}\|\hat{X}-\hat{U}\textsc{v})\|_{F}^{2}+\beta_{1}\|\textsc{v}\|_{2}^2
\end{equation}
where $X\in R^{m}$ is the current image and $\hat{X}=\mathcal P_{S}(X)$. By fixing $\hat{U}$, (\ref{eq:6}) is a least squares problem and can be solved by
\begin{equation}
\label{eq:update_V}
\begin{split}
\hat{\textsc{v}}=(\hat{U}^{T}\hat{U})^{\dagger}\hat{U}^{T}\hat{X} \,\,\,\,\,
\end{split}
\end{equation}
where $(.)^{\dagger}$ is the Moore Penrose pseudoinverse~\cite{bilateral_sketch}.

{\it{Part~2}}: To update $\hat{U}$,~(\ref{eq:6_0}) can be rewritten as:
\begin{equation}
\label{eq:6_1}
\begin{split}
\underset{\hat{U}} {\mathrm{min}}~\frac{1}{2}\|\hat{X}-\hat{U}\textsc{v}\|_{F}^{2}+ \beta_{1}\|\hat{U}\|_{F}^{2}\,\,\,\,\,
\end{split}
\end{equation}
and, according to Frobenius norm properties,~(\ref{eq:6_1}) can be solved by:
\begin{equation}
\label{eq:8}
\hat{U}=\underset{\hat{U}} {\mathrm{argmin}}~\frac{1}{2}Tr[\hat{U}(A+\beta_{1}I)\hat{U}^{T}]-Tr(\hat{U}^{T}\hat{B})
\end{equation}
where $A=\hat{\textsc{v}}\hat{\textsc{v}}^{T}$ and $\hat{B}=\hat{X}\hat{\textsc{v}}^T$. $\hat{U}$ means we update $U$ for those pixels that have foreground mask $s_i=1$.
%\begin{equation}
%\label{eq:6_2}
%\begin{split}
%\bar{U}=(\bar{X}-\bar{U}\textsc{v})\textsc{v}^T+2\beta_{1}\bar{U}
%\end{split}
%\end{equation}
Since $U$ is the basis of background for all images, it cannot be computed independently. This constraint of updating for $A$ and $B$ has been dealt with in~\cite{Related_OR-PCA}, where the basis $U$ minimizes a cumulative loss w.r.t the previously estimated coefficients $\textsc{v}$. Therefore, we use the following cumulative form to update $A$ and $B$, before computing $\hat{U}$ for the first iteration.
%We use the same approach as in~\cite{Related_OR-PCA} by defining two auxiliary parameters $A \in R^{r \times r}$ and $B \in R^{m \times r}$ in which to keep the information about $V$ and $U$ as follows:
\begin{equation}
\label{eq:7}
\begin{split}
A= A +\hat{\textsc{v}}\hat{\textsc{v}}^{T}\,\,\,\,\,\,\,\,\,\,\,\,\,\,\\
\hat{B}=\hat{B}+\hat{X}\hat{\textsc{v}}^{T}\,\,\,\,\,\,\,\,\,\,
\end{split}
\end{equation}
These accumulative forms enable us to use the previous background models to compute the current $U$ and keep the background model more stable against unexpected changes by increasing the number of images through time. In contrast to~\cite{Related_OR-PCA} we update $B$, only for those pixels that have foreground support $s_i=1$. Therefore, the number of rows in $\hat{B}$ is variable and equal to $|\textsc{s}|_0$ in each iteration.
In this part additive $A$ and $B$ save all previous information of $U$ and $\textsc{v}$ and are updated for the current image. By increasing the values of $A$ and $B$, the obtained background model becomes stable.

%{\it{Part~3}}: Based on the obtained $V$, $A$ and $B$, we update $U$ for the current image as follows~\cite{Related_OR-PCA}:

For the first iteration, $s_i=1$ for all pixels of the current image and so the number of rows in $\hat{B}$ and $\hat{U}$ is the same as that in the input image; subsequently, the number of rows in $\hat{B}$ and $\hat{U}$ decreases in succeeding iterations as the foreground area is decreased.~(\ref{eq:8}) can be solved with a simple iterative algorithm presented in~\cite{Related_OR-PCA}. Since COROLA is an iterative algorithm based on~(\ref{eq:3}) and the size of $\hat{U}$ and $\hat{B}$ changes in each iteration, in this implementation we save their values of $^{1}\hat{U}$, $^{1}\hat{\textsc{v}}$, $^{1}A$, and $^{1}\hat{B}$ after the first iteration. We use these values in the first iteration of the next input image.
Also these variables have the most information for building the background model of the current image, which is computed by $L=$$^{1}\hat{U}$$^{1}\hat{\textsc{v}}$. However, foreground detection depends on the obtained mask $\textsc{s}$ from the second step of solving~(\ref{eq:3}), and the algorithm continues to iterate until the convergence criteria are met. Because for dynamic backgrounds, outliers are a combination of  the foreground object and moving parts of the background as noise (e.g., waving trees). These moving parts do not affect background model, but they create false positives in the foreground mask $\textsc{s}$. We will explain the convergence criteria after solving the second step of (\ref{eq:3}).

\subsection{Online Foreground Detection}
\label{subsection3_3}
Let current $X$ and its corresponding $L$ be $X_{j}$ and $L_{j}$, respectively. Also $S_{j}$ is the indicator vector $\textbf{\textsc{s}}$ for the $j^{th}$ image. Now we investigate how to compute the foreground mask $\textbf{\textsc{s}}$ given the residual $E_{j}=X_{j}-L_{j}$ ($L_{j}$ is computed in background modeling in the previous section for the $j^{th}$ observed image). The goal now is to find the indicator vector $S_{j}$ on $E_{j}$. Assuming that the foreground objects are relatively small connected components, we can model the foreground mask $S_{j}$ by a Markov Random Field (MRF)~\cite{MRF} . Specifically, let graph $\mathcal G=(\mathcal V, \mathcal E)$ where $\mathcal V$ is the set of vertices that correspond to the pixels of an image and $\mathcal{E}$ is the set of edges that connect neighboring pixels. Then, by defining an energy function of $S_{j}$\begin{equation}
\label{eq:MRF}
\underset{i \in \mathcal V}\sum \beta_{2}(s_{i}) + \underset{(i,k)\in \mathcal E}\sum \gamma_{i,k}|s_{i}-s_{k}|
\end{equation}
which is called ``Icing model" in the literature and an example of MRF, we can derive the foreground mask $S_{j}$. The first and the second terms impose sparsity and continuity on $S_{j}$, in a way that is similar to the last two terms of~(\ref{eq:3}) and shows that $S_{j}$ can be modeled using MRF~\cite{MRF}.
 %and the energy of $S$ can be modeled as follows~\cite{MRF}.
%From the previous part, residuals $E_{j}=X_{j}-L_{j}$ are combination of outliers and noise.
However, extracting foreground objects from $E$, which is combination of outliers and noise, would not be accurate especially in noisy environment like dynamic backgrounds or with a moving camera. In most cases we need to separate reliable outliers representing true foreground from noise in estimating foreground support $S_{j}$. In most applications, noise comes from a complicated and dynamic background such as waving trees or sea waves, which should be classified as background.

\begin{figure}[t]
\centering
\includegraphics[scale=0.65]{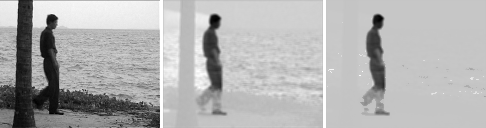}
\caption{\footnotesize{The effects of using GMM on outliers obtained from low rank approximation on noisy and dynamic background. The left figure shows an input image, and the middle and right figures show the obtained outliers $E$ and $\hat{E}$, respectively.}}
\label{fig:GMM1}
\end{figure}

Here, we describe outliers with a Gaussian model $\mathcal{N}(\mu,\sigma^{2})$. Using this model of the outliers enables us to control the complexity of the background variations and also recognize true outliers in the presence of noise using~(\ref{eq:9}). In our study, adaptive Gaussian Mixture Model (GMM)~\cite{AGMM} is used for each component of $E$ to separate the outliers from noise. As in most cases, three Gaussian components are sufficient in modeling $E$ to separate foreground $F$ from noise~\cite{AGMM}. Fig.~\ref{fig:GMM1} shows the effect of using GMM on $E$ for dynamic backgrounds.
The middle figure shows the obtained residual $E$. After obtaining $E$, we normalize it and extract outliers $F$ from noise using Gaussian model (right figure).
So, to solve the second step of~(\ref{eq:3}), we construct $\hat{E}$ with a simple update rule as follows:
\begin{equation}
\label{eq:9}
\hat{E}_{j}=\alpha E_{j}+(1-\alpha)F_{j}
\end{equation}
where $E_{j}=X_{j}-L_{j}$ and $F_{j}$ is the outliers using GMM on the current image ($j^{th}$ image of the sequence). $\alpha \in[0,1]$ is a constant that controls the magnitude of noise so that a small $\alpha$ would be used for noisy data (i.e. for moving cameras). In all of our experiments $\alpha=0.1$.

Now we can solve the second step of our optimization problem that extracts moving objects from outliers, and~(\ref{eq:3}) can be rewritten as the following objective function to minimize the energy over $S_{j}$ via obtained outliers $\hat{E}$.
\begin{equation}
\label{eq:10}
\begin{split}
\underset{S}{\mathrm{min}}~\frac{1}{2}\|\mathcal P_{S}(\hat{E})\|_{F}^{2}+\beta_{2}\|S_j\|_{1}+\gamma\|\Phi(S_j)\|_{1}+C\\
=\underset{i:s_{i}=1}{\sum}\hat{E}_{i}^{2}+\beta_{2}\underset{i}{\sum}s_{i}+\gamma\|\Phi(S_j)\|_{1}+C \,\,\,\,\, \\
\end{split}
\end{equation}
where $C$ is a constant. The first term of (\ref{eq:10}) is constant and therefore (\ref{eq:10}) is the first order MRF with binary labels (the same as (\ref{eq:MRF})), which can be solved using graph-cut~\cite{graph_cut1},~\cite{graph_cut2}. The result of~(\ref{eq:10}) is the binary mask $S_{j}$, which indicates the foreground pixels of $X_{j}$. So far, the first iteration of~(\ref{eq:3}) is completed and, based on mask $S_{j}$, the next iteration starts from~(\ref{eq:6}).
In our experiments, COROLA converges in approximately $r$ iterations where $r$ is the rank of data in the sequence. Our convergence criterion is similar to~\cite{Related_decolor} and we use $(energy_{prev}-energy)/energy < 10^{-4}$, where $energy=\frac{1}{2}\|(X_{j}-U\textsc{v})\|_{F}^{2}+\beta_{2}\|S_{j}\|_{1}$. In this formulation, the first and the second terms show the error of background model, and the foreground object size. The algorithm is considered to have converged if the error of background model and the size of the foreground object stabilize.
In Algorithm~\ref{alg:static}, we summarize all steps of COROLA.
\newcommand{\INDSTATE}[1][1]{\STATE\hspace{#1\algorithmicindent}}
\begin{algorithm}
\small
\caption{Online Moving Object Detection by COROLA}
  \label{alg:static}
  \begin{algorithmic}[1]
    \STATE Initialize: GMM parameters, $\beta_{1}$, $\beta_{2}$, $\gamma$, $\alpha_{1}$, $r$, $A$, and $B$
    \STATE for $j=1:n$
    \INDSTATE[1] Input data: $X_{j}$
    \INDSTATE[1] $t=1$ and  $s_{i}=1$,  $i=\{1,...,m\}$
    \INDSTATE[1] repeat
    \INDSTATE[2] If  $t=1$
    \INDSTATE[3] $\hat{A} \longleftarrow A_{j-1}$, $\hat{B} \longleftarrow B_{j-1}$ , $\hat{U} \longleftarrow U_{j-1}$
    \INDSTATE[2] else
    \INDSTATE[3] $\hat{A} \longleftarrow A_{j}$, $\hat{B} \longleftarrow B_{j}$, $\hat{U} \longleftarrow U_{j}$
    \INDSTATE[2] end If
    \INDSTATE[2] $\hat{V}_{j}\longleftarrow \underset{V_j} {\mathrm{argmin}}~\frac{1}{2}\|\hat{X}_{j}-\hat{U}V_{j}\|_{F}^{2}+\beta_{1}\|V_j\|_{2}^2$, where $\hat{X}_j =\mathcal P_{S}(X_j), \hat{U}=\mathcal P_{S,:}(U)$
    \INDSTATE[2] $A_{j}\longleftarrow \hat{A}+\hat{V}_{j}\hat{V}_{j}^{T}, \,\,\, \hat{B}_{j} \longleftarrow \hat{B}_{j-1}+ \hat{X}_{j}\hat{V}_{j}^{T}$
    \INDSTATE[2] $\hat{U}_{j} \longleftarrow \underset{\hat{U}} {\mathrm{argmin}}~Tr[\hat{U}(A_{j}+\beta_{1}I)\hat{U}^{T}]-Tr(\hat{U}^{T}(B_{j}))$
    \INDSTATE[2] $E_{j} \longleftarrow X_{j}-L_{j}$, \,\,\, compute $F_{j}$ over $E_{j}$ from~\cite{AGMM}
    \INDSTATE[2] $\hat{E}_{j}\longleftarrow \alpha E_{j}+(1-\alpha)F_{j}$
    \INDSTATE[2] Compute cost of assigning labels using $\hat{E}_{j}$ to optimize $S$
    \INDSTATE[2] $\textsc{s} \longleftarrow \underset{S} {\mathrm{argmin}} \,\, \beta_{2}\underset{i}{\sum}s_{i}+\gamma\|\Phi(S_j)\|_{1}$
    \INDSTATE[2] If $t=1$,
    \INDSTATE[3] $^{1}\hat{U}_{j} \leftarrow \hat{U}_{j}$, $^{1}\hat{V}_{j} \leftarrow \hat{V}_{j}$, $^{1}A_{j} \leftarrow A_{j}$, and $^{1}B_{j} \leftarrow B_{j}$
    \INDSTATE[2] end If
    \INDSTATE[2] If $t \geq r$
    \INDSTATE[3] break
    \INDSTATE[2] else
    \INDSTATE[3] $t \longleftarrow t + 1$
    \INDSTATE[2] end If
    \INDSTATE[1] until convergence
    \INDSTATE[1] Output: $S_{j}$, $L_{j}=$$^{1}\hat{U}_{j}$$^{1}\hat{V}_{j}$
    \INDSTATE  $\hat{U}_{j} \leftarrow$ $^{1}\hat{U}_{j}$, $\hat{V}_{j} \leftarrow$ $^{1}\hat{V}_{j}$, $A_{j} \leftarrow$ $^{1}A_{j}$, and $B_{j} \leftarrow$ $^{1}B_{j}$
    \STATE end for
  \end{algorithmic}
\end{algorithm}
\subsection{Convergence of COROLA}
In this section, we explain the convergence criteria of COROLA. In general, our main objective function~\ref{eq:3} is non-convex and we solve it by alternating between two steps. In step one for low-rank approximation, we always minimize a single lower-bounded energy function using OR-PCA. The convergence propoerty of OR-PCA has been proved in~\cite{Related_OR-PCA}. In the second step for outlier detection, we use MRF and its convergence has been discussed in~\cite{graph_cut1}. Using these two steps, the algorithm must converge to a local minimum; furthermore,~\cite{Related_decolor} showed that this combinatorial optimization decreases the energy monotonically through iterations and can converge to acceptable results in background modelling and moving object applications.

\subsection{Online Moving Object Detection with a Moving Camera}
\label{subsection3_4}
In this part, we extend our moving object detection method to the case of a moving camera. As we mentioned in Section~\ref{section1}, due to the dissimilarity between the first and the last images in a sequence, a batch method is not able to deal with continuous processing using a moving camera. However, in online methods the background model evolves with time and similarity between the first and the current image is not required. In  our method, we build the background model for the current image and based on a transformation function between the current and the new image, the model is transformed to be matched with the new image. Then we can update it for the new image to detect the foreground objects. Note that the background model is transformed through time. So the key in foreground detection using a moving camera is the transformation of the low-rank structure to the new input image.

%To show the capability of our method in comparison with the DECOLOR method, we use the same image transformation parameters as DECOLOR.
Let $\tau_{j}$ be a transformation that maps $X_{j-1}$ to $X_{j}$. This transformation is obtained from an affine transformation estimated from the two 2D images. We also assume $X_{j-1}={U}_{j-1}{\textsc{v}}_{j-1}$ and there is no changes into both images except for affine transformation so that $X_{j} = \tau \circ X_{j-1}$.
For the sake of brevity, we state without proof that the following equation allows us to reconstruct the current view $X_j$ from the background model and the registration transform $\tau_j$.
\begin{equation}
\label{eq:move1}
X_{j} = \tau_j \circ X_{j-1} = (\tau_{j} \circ {U}_{j-1}){\textsc{v}}_{j-1}
\end{equation}
From~(\ref{eq:move1}) the transformation only changes ${U}$. In fact, we need to transform $B$ via $\tau$ only once for the first iteration of each input image
%\begin{equation}
%\label{eq:move2}
%\mathcal P_{S,:}(B_{j}) = \mathcal P_{S,:}(\tau_{j} \circ {B}_{j-1}+\bar{U}_{j-1}\hat{V}_{j}\hat{V}_{j}^{T})
%\end{equation}
where $\bar{U}_{j-1}=\tau \circ U_{j-1}$ and $\bar{B}_{j-1}=\tau \circ B_{j-1}$. In~(\ref{eq:7}), $A$ remains unchanged, because $\hat{\textsc{v}}$ is independent from $\tau$. Based on the above assumptions and~(\ref{eq:move1}), $U_{j} = \bar{U}_{j-1}$ and $\textsc{v}_j={\textsc{v}}_{j-1}$.
After the transformation, some elements of $\bar{U}_{j-1}$ and $\bar{B}_{j-1}$, which are related to the pixels on the border of the current image, have no corresponding pixels and we have to estimate them using other pixels. To solve the problem, first we normalize both $\bar{U}_{j-1}$ and the current image to [0,1]. Then, using $X_j$ and $\textsc{v}_{j-1}$ (or $\textsc{v}_j$) we estimate missing pixels of $\bar{U}_{j-1}$ by replacing them by the corresponding values obtained from~\cite{bilateral_sketch} and ensure they lie in the correct range, as follows.
\begin{equation}
\label{eq:moving_U}
\bar{U}_{j-1}=X_j\textsc{v}^T_{j-1}(\textsc{v}_{j-1}\textsc{v}_{j-1}^T)^{\dagger}\,\,\,\,\,
\end{equation}
Similarly, for estimating missing pixels of transformed $\bar{B}_{j-1}$, we normalize both $\bar{B}_{j-1}$ and $\bar{U}_{j-1}\textsc{v}_{j}\textsc{v}_{j}^T$ and we replace those missing values of $\bar{B}_{j-1}$ with $\bar{U}_{j-1}\textsc{v}_{j}\textsc{v}_{j}^T$ (from~(\ref{eq:7})) and ensure they lie in the correct range.

\begin{figure}[t]
\centering
\includegraphics[scale=0.9]{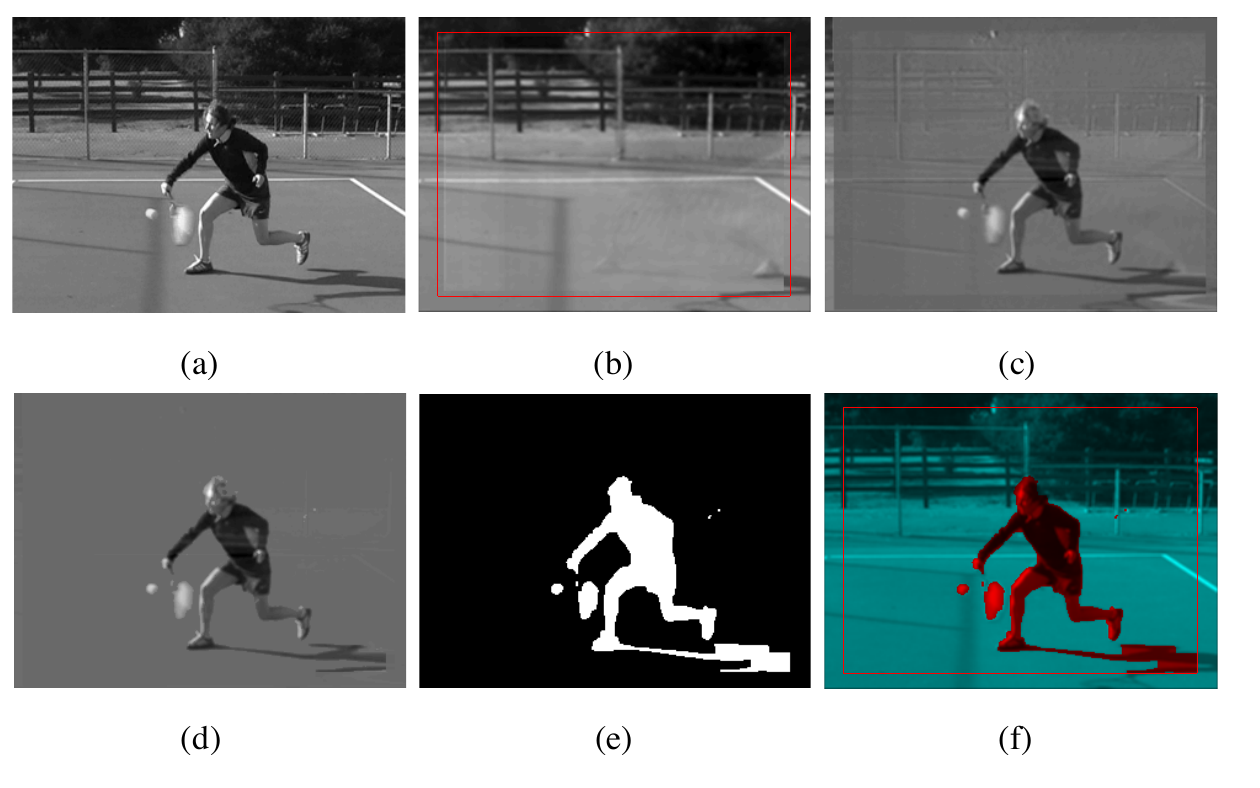}
\caption{\footnotesize{An example of COROLA for a moving camera. (a) input image from a sequence (b) background model (c) $E$, (d) $\hat{E}$, (e) $S$, and (f) extracted foreground object using mask $S$. Red lines show the processing area.}}
\label{fig:moving}
\end{figure}

Based on the experimental results, this approach can estimate missing pixels of $U$ and $B$ after transformation. In addition, the GMM for the previous $E_{j-1}$ should be transformed via $\tau$ to match with the current $E_{j}$. After transforming $U$,$B$, we can apply the COROLA method for a static camera to build the background model and detect the foreground objects. Fig.~\ref{fig:moving} shows a sample image, its computed background model and extracted moving object via COROLA, together with the intermediate results.

%\subsection{Time Complexity}
%\label{subsection3_5}
%In the next section, we compare the computational cost of  COROLA and DECOLOR.
The complexity of our sequential low-rank approximation by COROLA consists of contributions from two major parts. The computational complexity of the first part is $O(mr)$. The second part of the low-rank approximation in our model are $O(r^2+mr)+O(mr^2)$. Therefore, the overall complexity of COROLA for the low-rank approximation step is $O(r^2+mr^2)$.

\section{Experimental Results}
\label{section4}
In this section, we compare COROLA with competing algorithms in the literature. We perform two sets of experiments on synthetic data and real benchmark datasets and show quantitative and qualitative results. For quantitative evaluation where ground truth is available, we use pixel-level precision and recall, defined as follows:
\begin{equation}
\label{eq:11}
precision=\frac{TP}{TP+FP},\,\,\,\,\,recall=\frac{TP}{TP+FN}
\end{equation}
where TP, FP, TN, and FN are the numbers of true positives, false positives, true negatives and false negatives, in pixels, respectively. Also, instead of using precision-recall curves, we use F-measure to show the overall accuracy.
\begin{equation}
\label{eq:12}
{\text{F-measure}} = 2~\frac{precision \times recall}{precision + recall}
\end{equation}
%Furthermore, In simulation part we use the Root Mean Square Error (RMSE) as follows to show the error of recovered background model.
%\begin{equation}
%\label{eq:13}
%RMSE=\frac{\|L-B\|_{F}}{\|B\|_{0}}
%\end{equation}
%where $B$ is the Ground-Truth (GT).
\subsection{Synthetic Data}
\label{subsection5_1}
In this set of experiments, we use synthetic data to control noise and to show the capability of COROLA. The synthesized images are $30 \times 100$ pixels ($m=3000$). We use $n=200$ images. Zhou {\it{et. al.}}~\cite{Related_decolor} used the similar scheme to investigate the robustness of their method against outliers.% and the size of moving object is $30 \times 8$ which is $240$ pixels.

\begin{figure}[!t]
\centering
\includegraphics[scale=0.9]{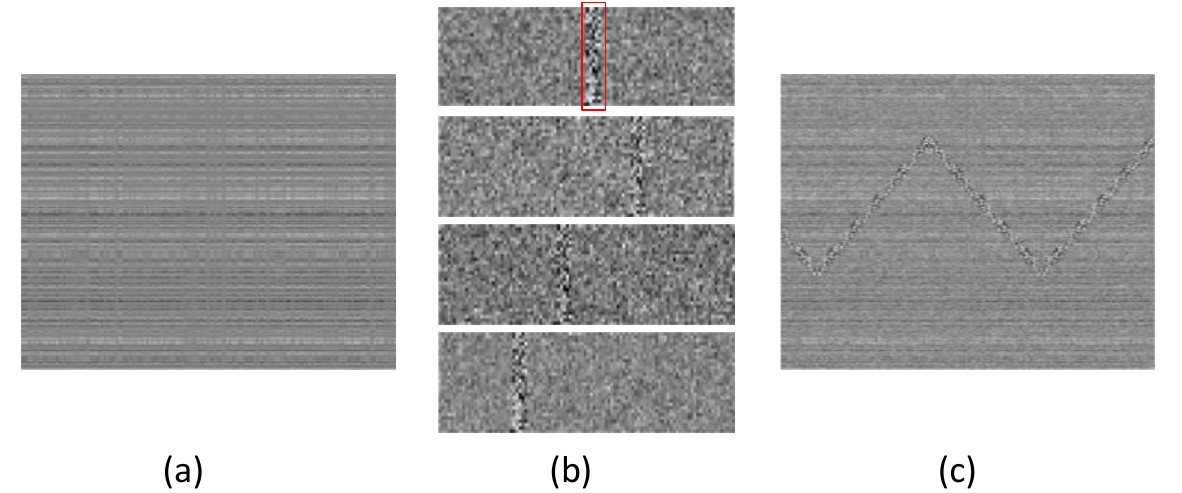}
\caption{\footnotesize{An example of synthetic data. (a) shows matrix $B \in R^{3000 \times 200}$, with $m=3000$, $n=200$, and rank $r=5$, where $B=UV$, $U \in R^{3000 \times 5}$, and $V \in R^{5 \times 200}$. (b) shows some sample images from selected column of $B$, where an object is superimposed each of them. The object is represented by a red box in the first image in (b). other images show the movement of the object to left and right of the image, frequently. (c) shows a sample of generated matrix $D$ as the input data. }}
\label{pics:SIMU_1}
\end{figure}

To visualize the results we show all images in a 2D matrix where each column shows one image of the sequence. We generate the input data $D$ by adding a foreground to a background matrix $B$. For generating the foreground and background we use the same approach as DECOLOR. The background matrix $B=UV$ is generated via $U \in R^{m \times r}$ and $V \in R^{r \times n}$ with random samples from a standard normal distribution. An object with a small size is superimposed on each image in matrix $B$, and shifts from left to right of the images by one pixel per image, until the right border of the image. The motion direction of the object is then reversed, and the process repeats. Fig.~\ref{pics:SIMU_1}(b) shows some selected images. The intensity of this object is independently sampled from a uniform distribution.
Also, we add i.i.d Gaussian noise $\epsilon$ to $D$ with the corresponding signal-to-noise ratio defined as
\begin{equation}
\label{eq:14}
SNR=\sqrt{\frac{var(B)}{var(\epsilon)}}
\end{equation}
Figs.~\ref{pics:SIMU_1}(a), (b) and (c) show an example of generated $B$, the movement of generated foregrounds and the obtained matrix $D$.

\begin{figure}[!b]
\begin{center}
\begin{tabular}{l}
\includegraphics[scale=0.6]{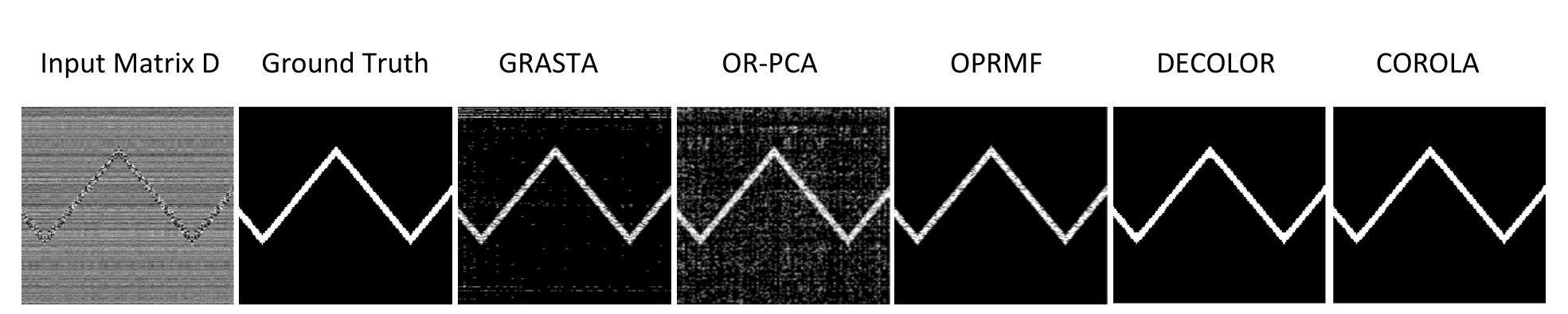} \\
\includegraphics[scale=0.6]{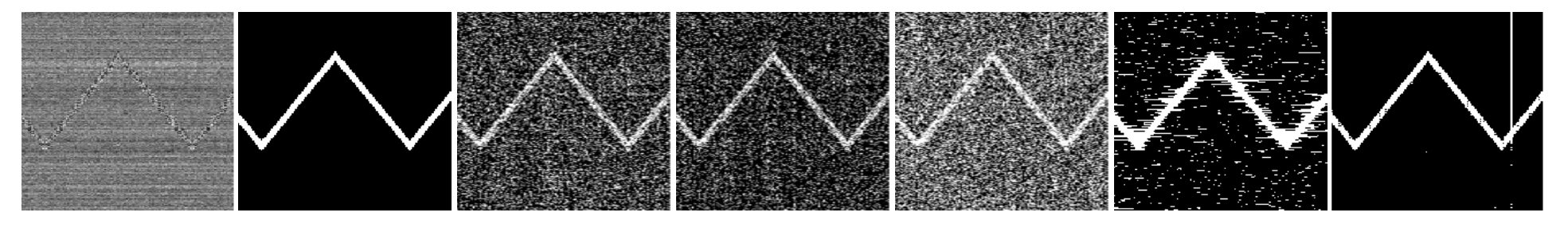}
\end{tabular}
\end{center}
\caption{\footnotesize{Comparison of COROLA, GRASTA, ORPCA, OPRMF and DECOLOR with different SNR ratio. The first row and the second row show the results of the methods with $SNR=10$, and $SNR=1$, respectively.}}
\label{pics:sample}
\end{figure}

We test the COROLA method and compare it with leading online methods such as GRASTA, OPRMF, ORPCA and DECOLOR, one of the best batch methods, in terms of different SNR ratios, different ranks of matrix, and different sizes of the foreground object. One sample of our experiments with different SNR ratios between COROLA and all mentioned methods is shown in Fig.~\ref{pics:sample}. In the first row of Fig.~\ref{pics:sample}, with $SNR=10$, COROLA, OPRMF and DECOLOR methods have roughly the same results for extracting the foreground object, but when we increase noise in the second row ($SNR=1$), COROLA method works better than all other methods including DECOLOR in extracting the moving object. That is mainly attributed to using GMM to compute the coefficients of outliers to separate the foreground object from noise. Tuning up the outliers coefficient via GMM enables us to separate noise and outliers especially in a noisy environment and the result becomes more and more accurate over time.

\begin{figure}[t]
\begin{center}$
\begin{array}{ccc}
\includegraphics[scale=0.25]{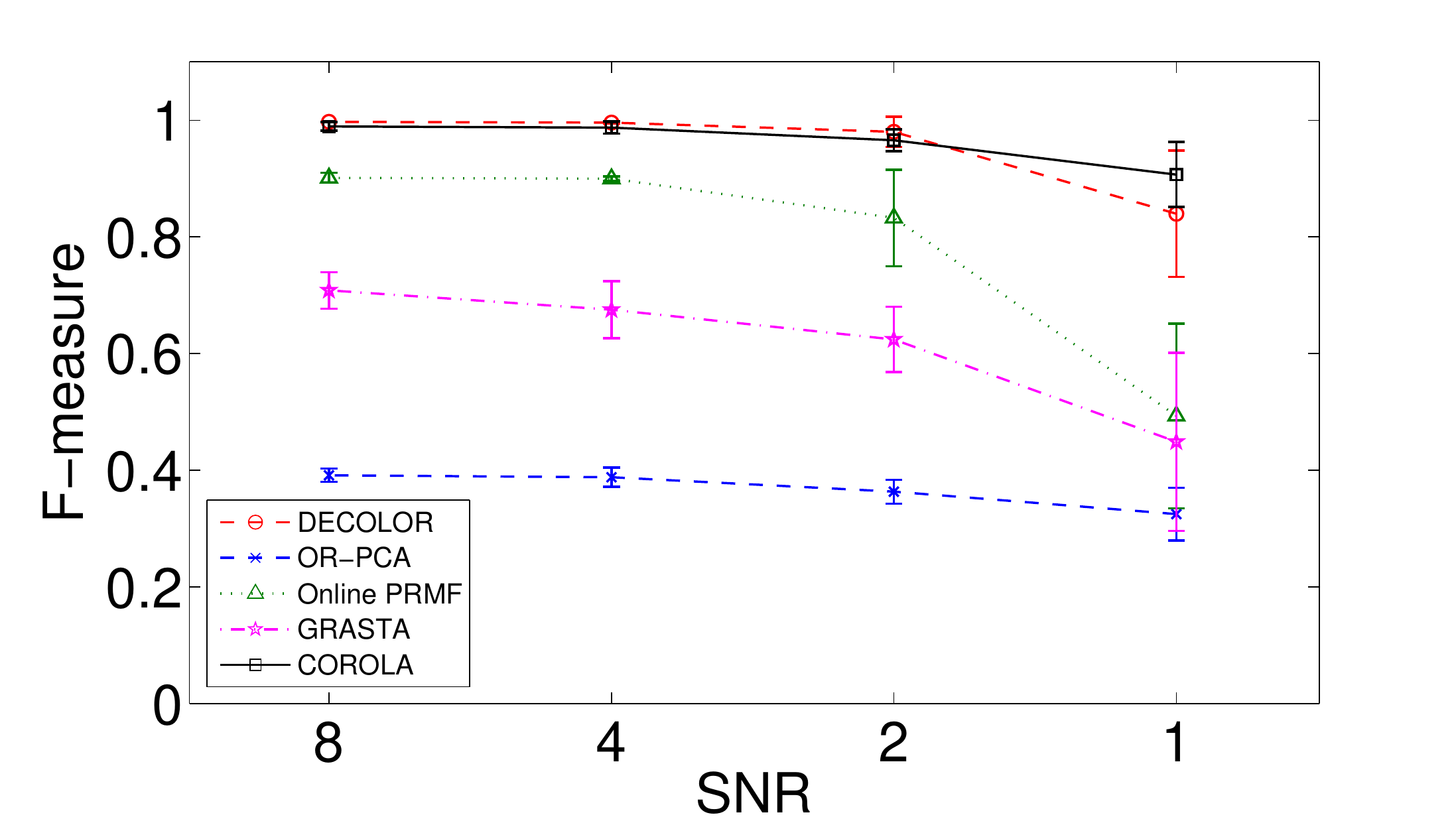}&
\includegraphics[scale=0.25]{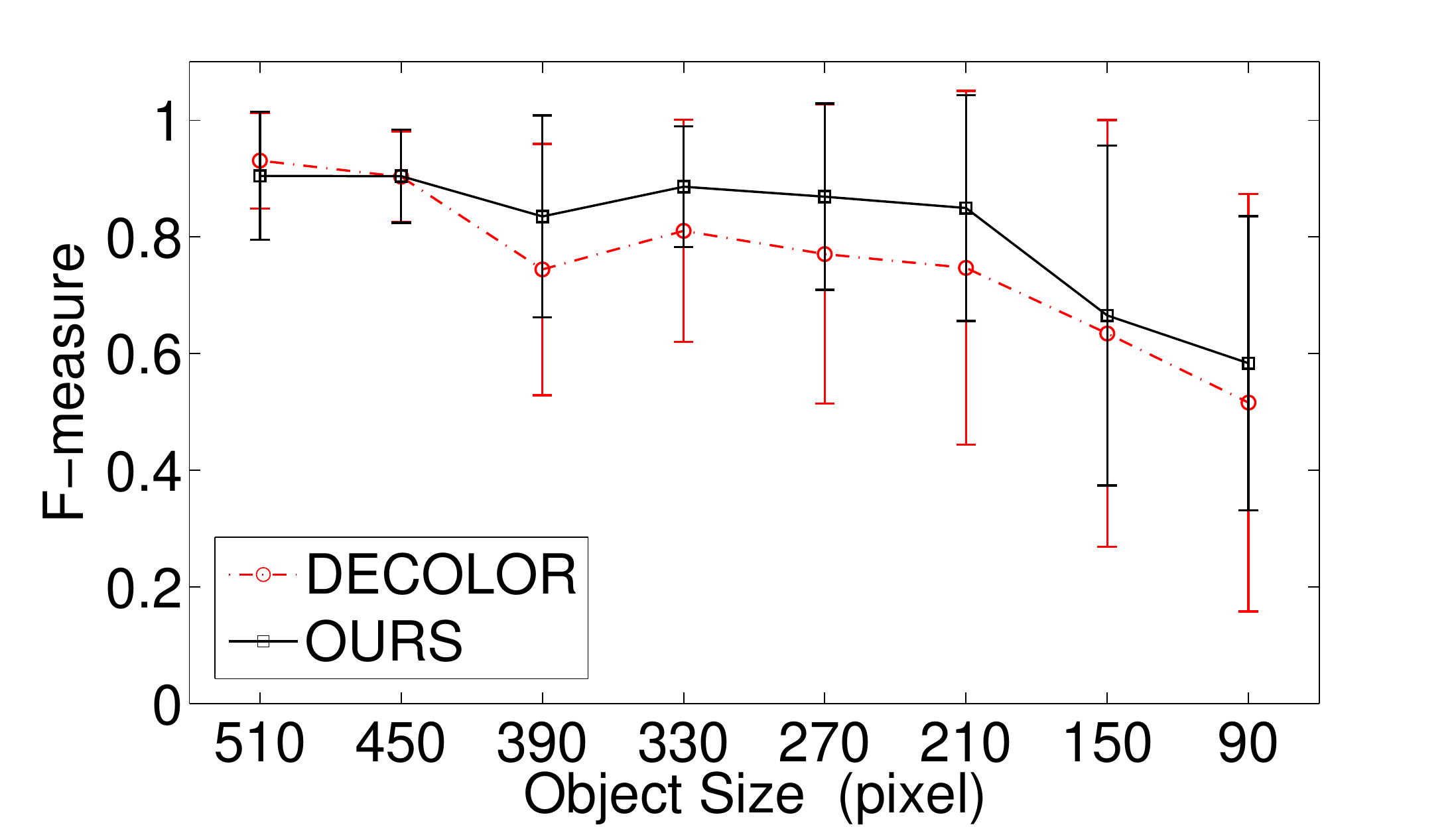}\\
\includegraphics[scale=0.25]{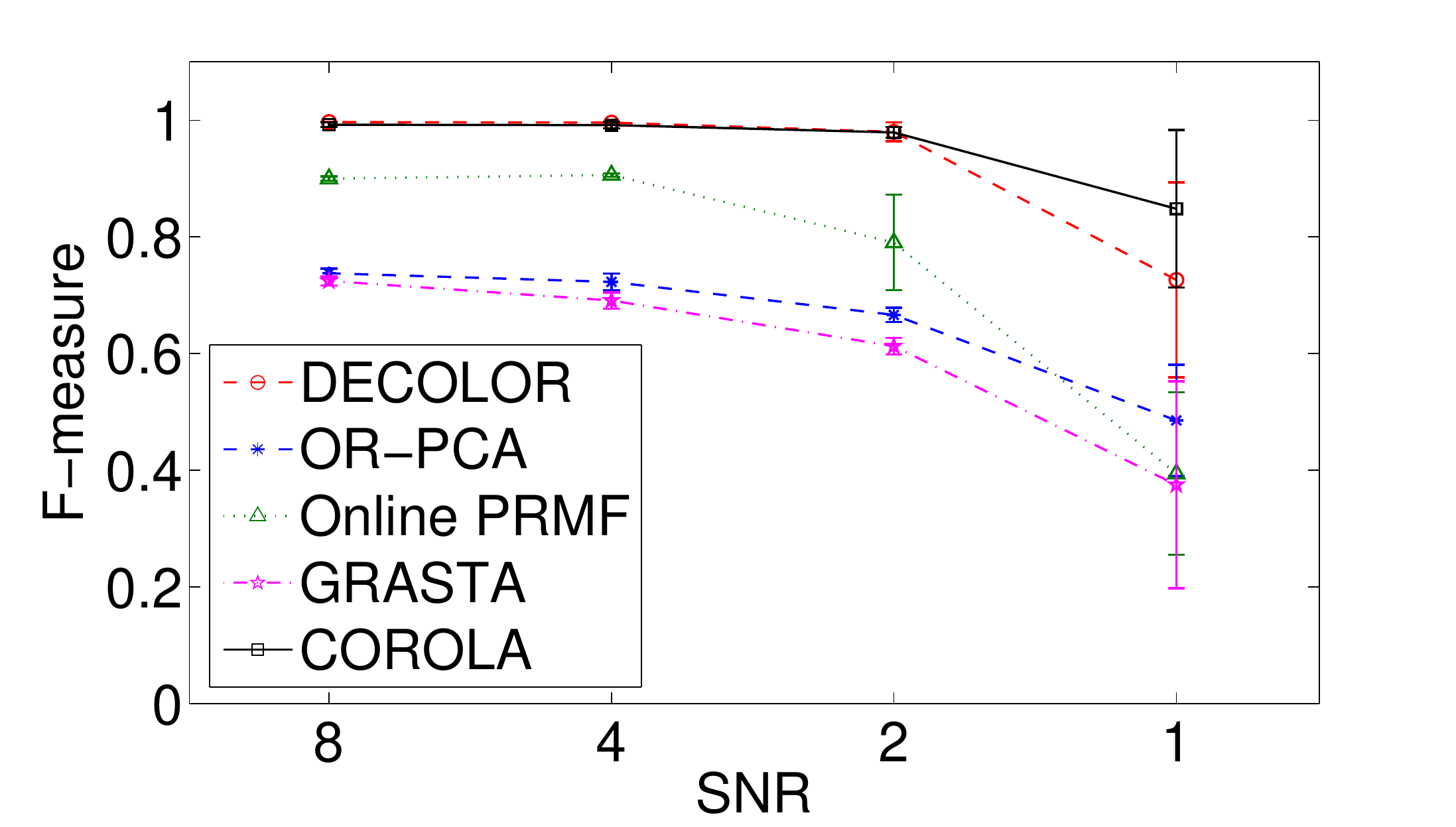}&
\includegraphics[scale=0.25]{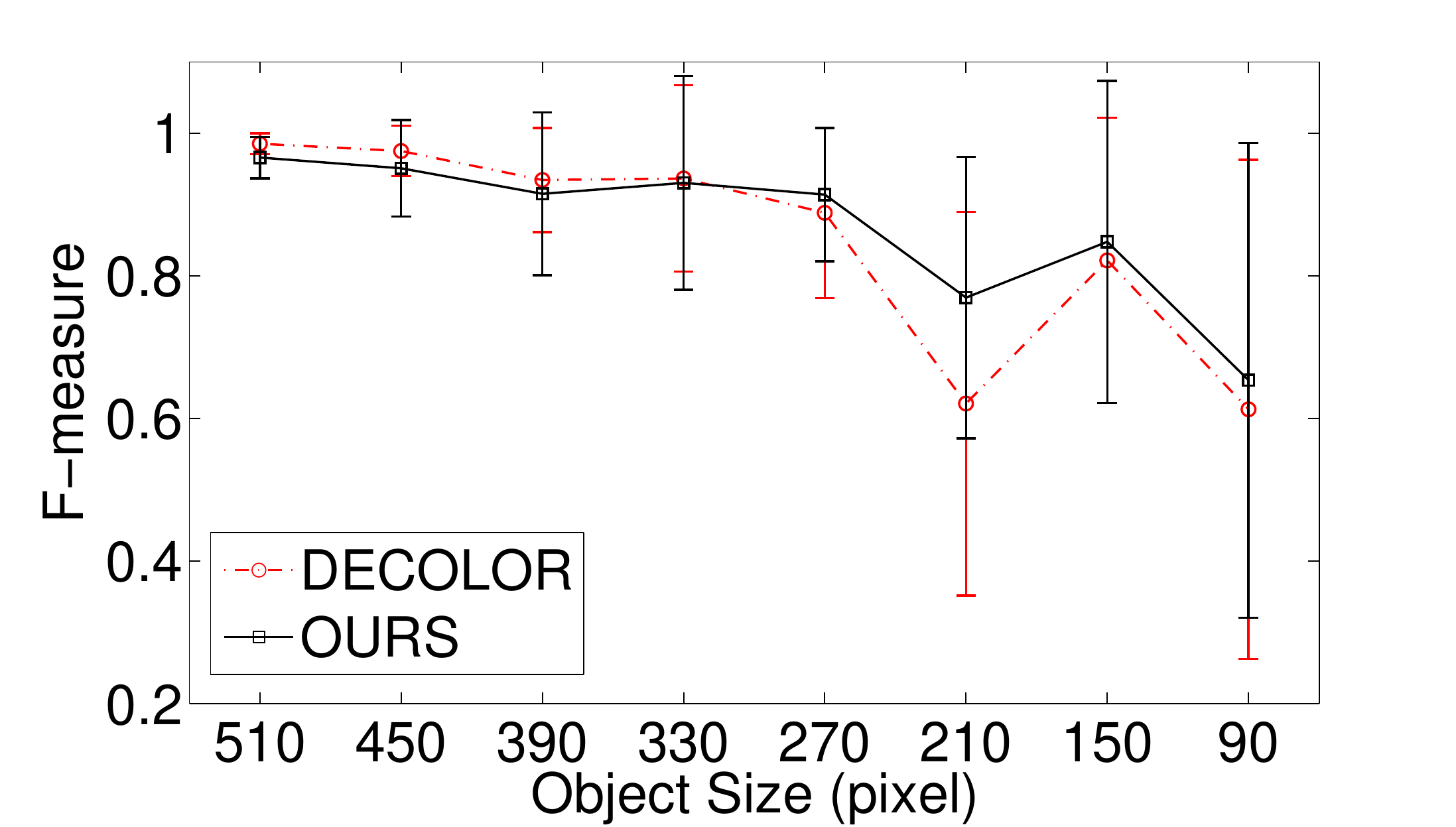}\\
\includegraphics[scale=0.25]{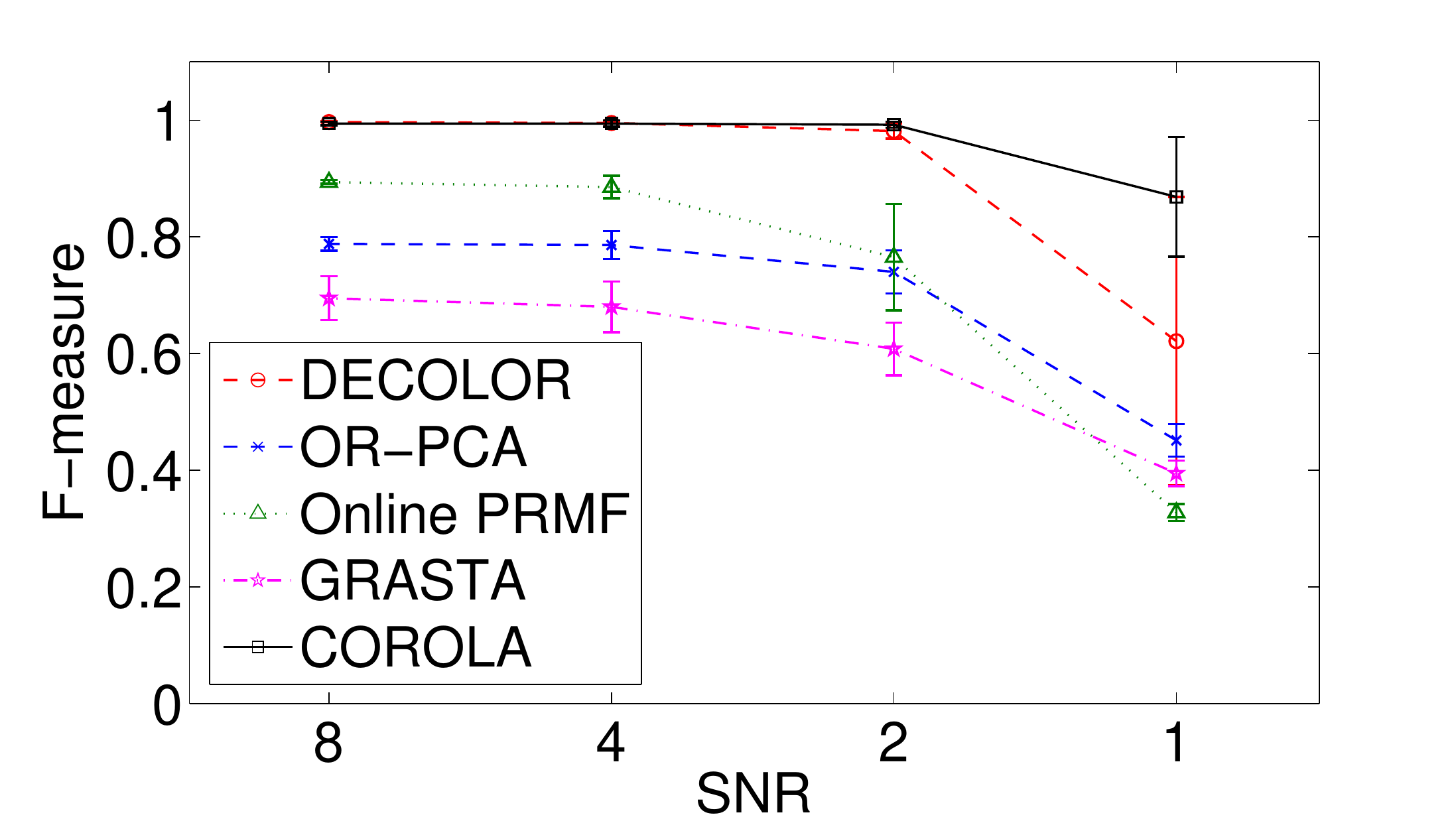}&
\includegraphics[scale=0.25]{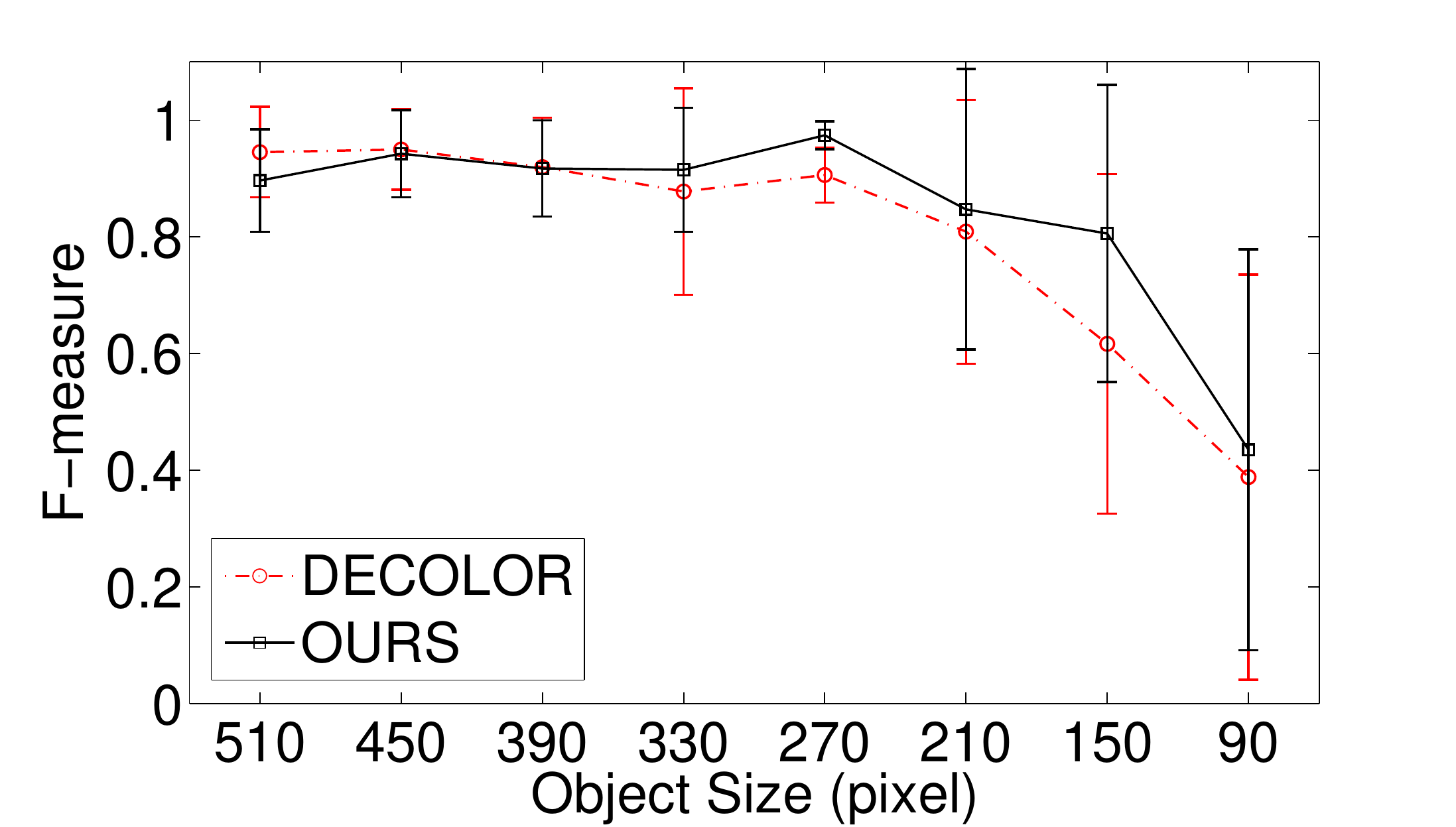}
\end{array}$
\end{center}
\caption{\footnotesize{First column: the comparison in terms of F-measure between COROLA and other methods with different signal-to-noise (SNR) ratio. Second column: the comparison of F-measure between COROLA and DECOLOR with different object size. The three rows show three different ranks at 1, 3, and 5 respectively.}}
\label{pics:SNR_objectsize}
\end{figure}

To evaluate COROLA in comparison with GRASTA, OPRMF, ORPCA, and DECOLOR methods, we tested the effects of some scene parameters such as SNR, rank of matrix $D$, and size of the object. The quantitative results of this comparison in terms of F-measure are provided in Fig.~\ref{pics:SNR_objectsize}. The first column of Fig.~\ref{pics:SNR_objectsize} illustrates the effect of noise in all methods, when we change the SNR ratio from 8 to 1 in different ranks. The rows from top to bottom show our experiments in different ranks of 1, 3, and 5. Since one of the advantages of DECOLOR method is high accuracy of object detection with different sizes, the second column of Fig.~\ref{pics:SNR_objectsize} shows the accuracy of COROLA in comparison with DECOLOR to extract the moving object of different sizes.
This result demonstrates that the capability of our method is comparable with DECOLOR in terms of average F-measure. Although, the result of DECOLOR method is slightly more accurate than COROLA for large objects, by reducing the size of object, COROLA generates a better result than DECOLOR even when we increase the rank of matrix $D$ from 1 to 5.

\begin{figure}[t]
\begin{center}$
\begin{array}{ccc}
\includegraphics[scale=0.19]{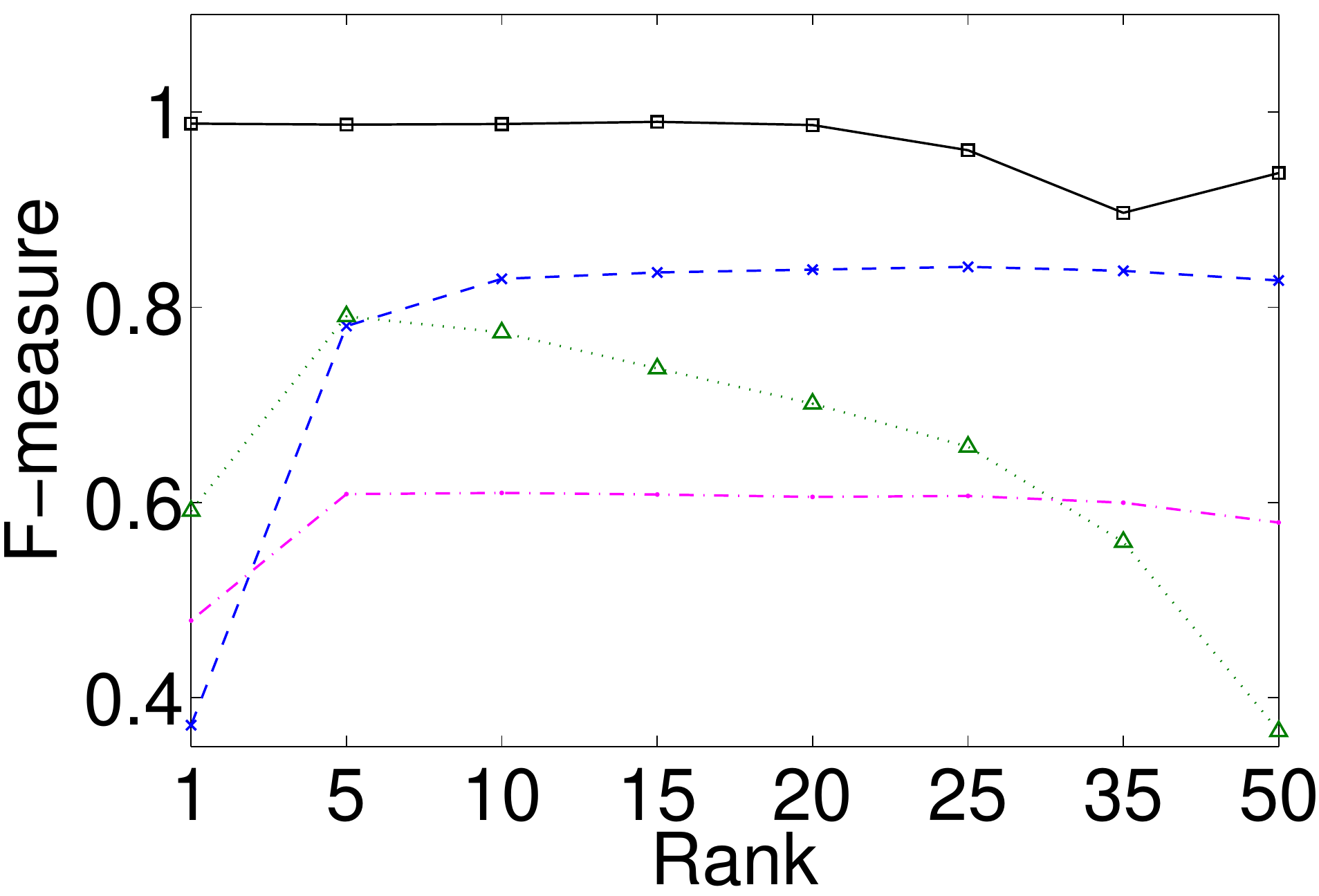}&
\includegraphics[scale=0.19]{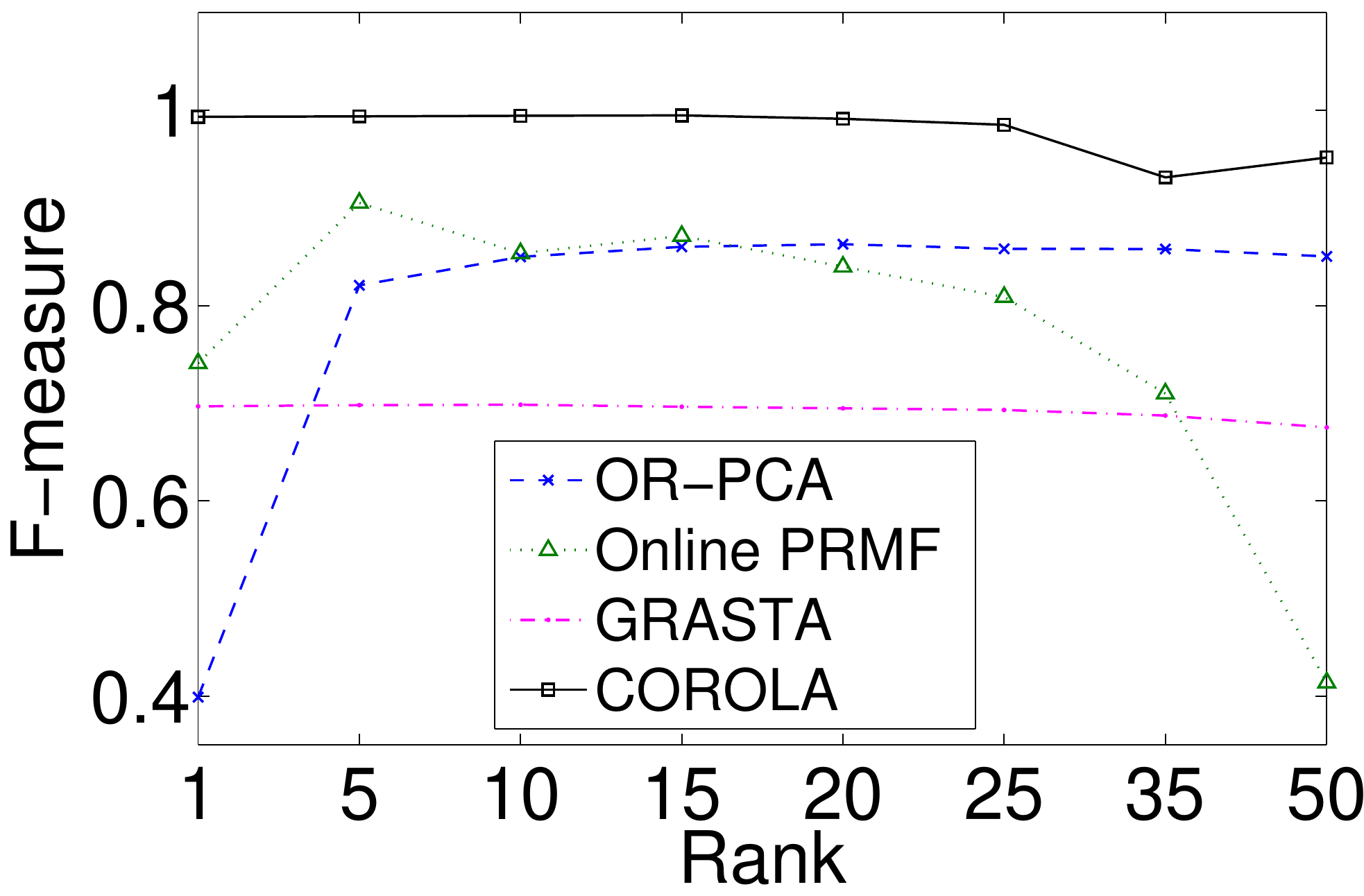}&
\includegraphics[scale=0.19]{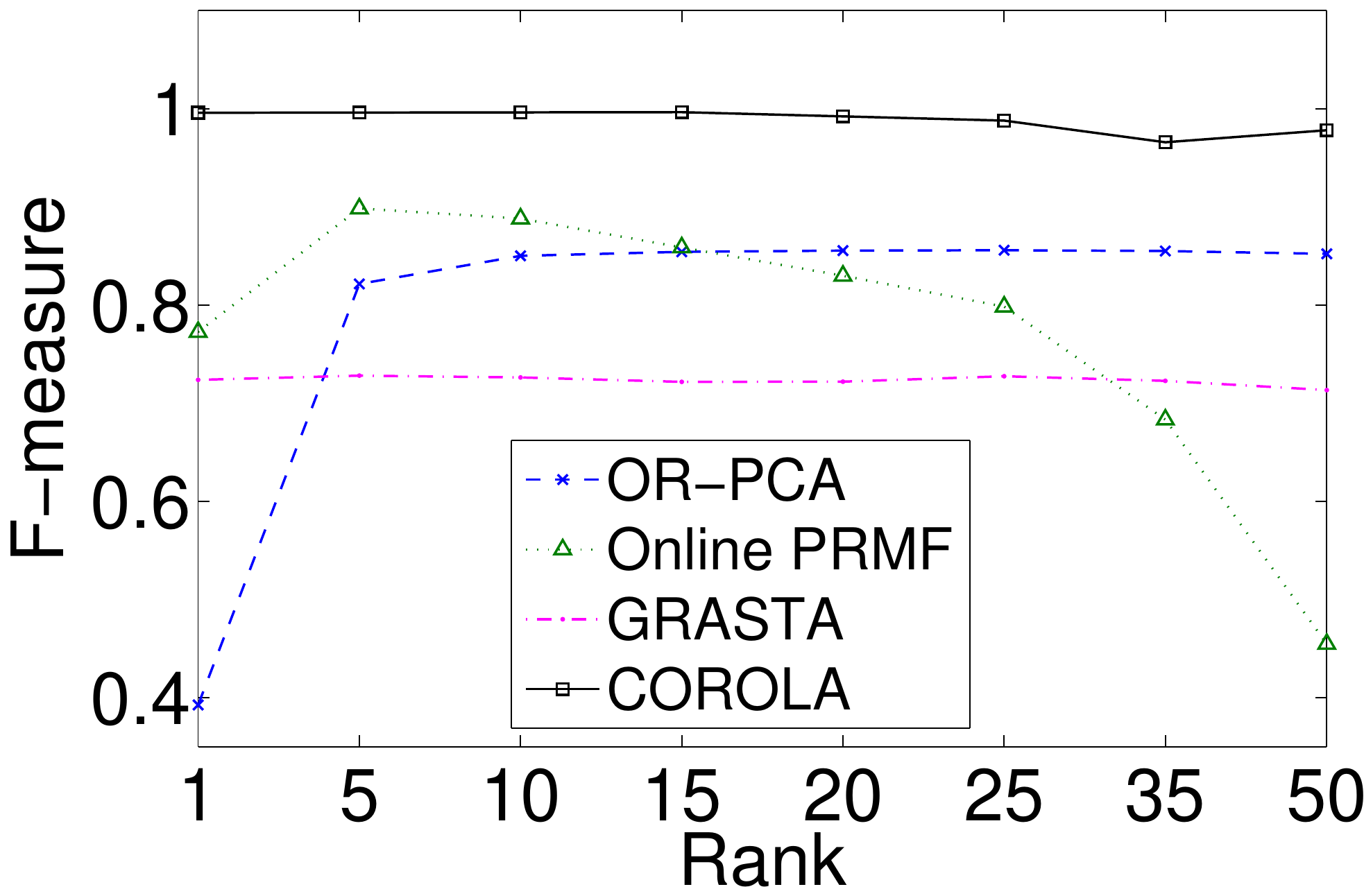}
\end{array}$
\end{center}
\caption{\footnotesize{Comparison of F-measure between COROLA and other methods in different ranks. True rank is 3, and columns show the results with different SNR.}}
\label{pics:Rank_sensitive}
\end{figure}
To evaluate the rank sensitivity of the COLORA method, we tested the effects of changing the rank of our method against other online methods. Fig~\ref{pics:Rank_sensitive} demonstrates F-measure of these methods, when we set the rank of the methods from 1 to 50.
The columns from left to right show our experiments in different SNR 2, 4, and 8 on the synthetic data with the true rank 3. In this experiment, when we set the rank of methods less than the true rank of the data, ORPCA, OPRMF, and GRASTA failed. This is because when the rank of data is higher than the predefined rank, these methods consider some of background variations as positive foreground pixels, incorrectly. In contrast, COROLA can extract foreground objects even with a lower rank than the true rank. It is because GMM allows COROLA to remove false positive pixels. In Fig.~\ref{pics:Rank_sensitive} by increasing the predefined rank of the methods, GRASTA, OR-PCA, and COROLA are robust although COROLA still shows the highest F-measure against all other methods.

\subsection{Real Data}
In this section, we use real benchmark datasets to conduct quantitative and qualitative evaluation of COROLA and compare it with DECOLOR, MOG, SSGoDec, and ORPCA+MRF. The real datasets used are popular in moving object detection and publicly available\footnote{https://sites.google.com/site/backgroundsubtraction/test-sequences}, and they include ``2014 Change Detection"~\cite{dataset_change}, ``Perception or I2R"~\cite{dataset_i2r}, and ``Wallflower"~\cite{dataset_wallflower} test images sequences. Table~\ref{table_static_dataset} provides the length and image size of these datasets.
\setlength{\tabcolsep}{3pt}
\begin{table}[hb]
\scriptsize
\begin{center}
\caption{\footnotesize{Details of all sequences used in our experiments for stationary camera}}
\label{table_static_dataset}
\begin{tabular}{lllllll}
\hline\noalign{\smallskip}
Dataset &&& Sequences &&& Size $\times$ \#frames\\
\cline{1-7}
I2R              &&& Water surface &&& [160,128] $\times$ 48  \\
                 &&& Fountain      &&& [160,128] $\times$ 523 \\
                 &&& Curtain       &&& [160,128] $\times$ 2964\\
                 &&& Hall          &&& [176,144] $\times$ 1927\\
                 &&& Campus        &&& [160,128] $\times$ 372 \\
                 &&& Escalator     &&& [160,130] $\times$ 824 \\
                 &&& Lobby         &&& [160,128] $\times$ 138 \\
                 &&& ShoppingMall  &&& [320,256] $\times$ 433 \\
\cline{1-7}
Change Detection &&& Canoe     &&& [320,240] $\times$ 1189\\
                 &&&Fall       &&& [180,120] $\times$ 1500\\
                 &&&Fountain02 &&& [216,144] $\times$ 720 \\
                 &&&Overpass   &&& [320,240] $\times$ 3000\\
\cline{1-7}
Wallflower       &&& Waving trees       &&& [160,120] $\times$ 287 \\
                 &&& Bootstrap          &&& [160,120] $\times$ 299 \\
                 &&& Camouflage         &&& [160,120] $\times$ 251 \\
                 &&& ForegroundAperture &&& [160,120] $\times$ 489 \\
 %                &&& LightSwitch        &&&                        \\
                 &&& TimeOfDay          &&& [160,120] $\times$ 1850\\
\cline{1-7}
%BMC &&& Boring parking &&& [240,320] $\times$ 1565\\
\hline
\end{tabular}
\end{center}
\end{table}
\setlength{\tabcolsep}{1.4pt}

\subsubsection{Evaluation by accuracy}

\begin{figure}[t]
\centering
\includegraphics[scale=0.6]{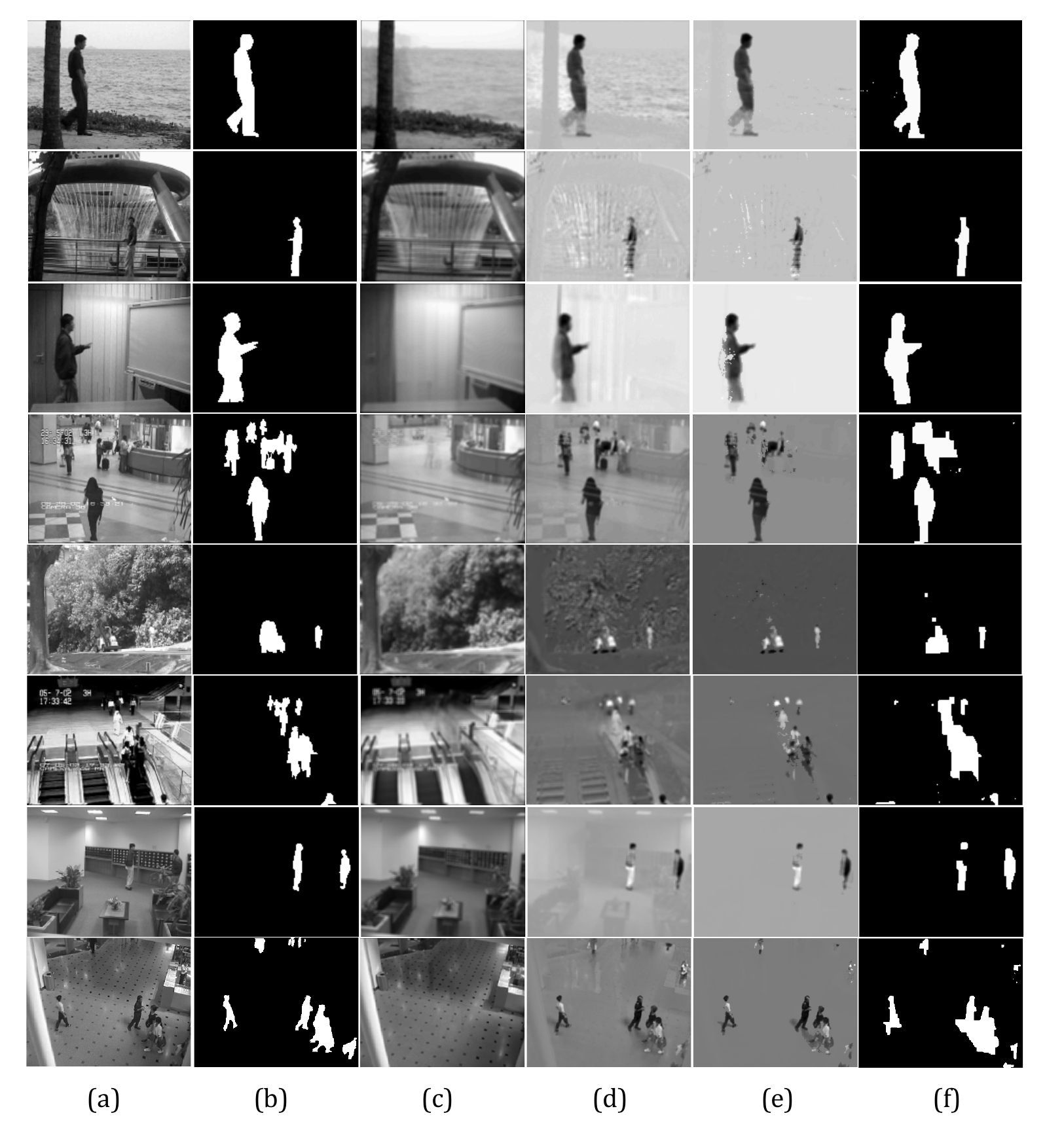}
\caption{\footnotesize{The results of COROLA on 6 sequences from three detasets Change Detection, I2R, and Wallflower. Columns (a) and (b) show the original query image and the ground truth (GT) for the foreground. Columns (c) and (f) show the results of COROLA for estimating the background $L$, and the detected foreground objects $S$, respectively. Columns (d) and (e) show intermediate results for outliers $E$, and $\hat{E}$, respectively. }}
\label{fig:qualitative_s1}
\end{figure}

\begin{figure}[t]
\centering
\includegraphics[scale=0.61]{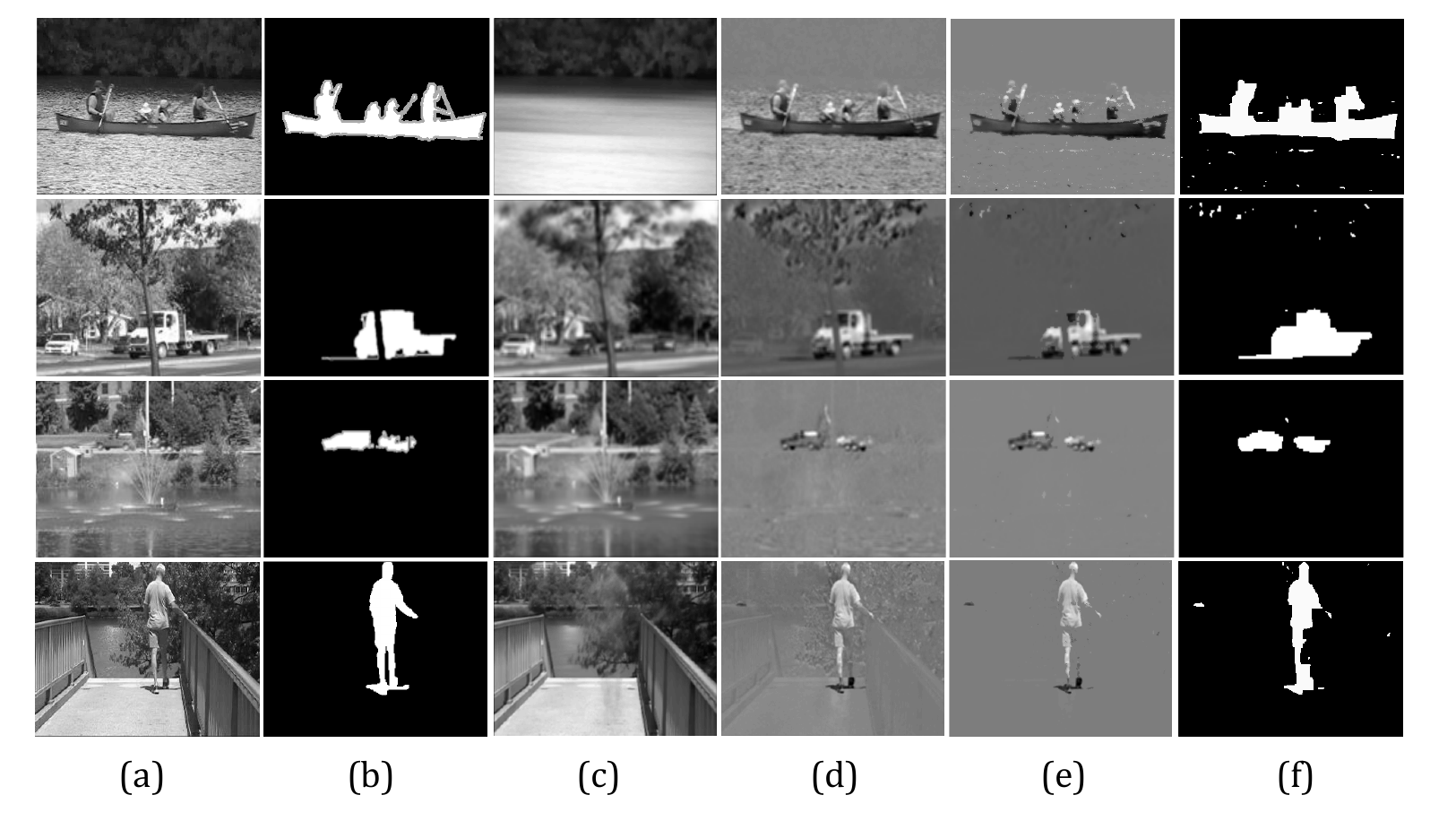}
\caption{\footnotesize{The results of COROLA on 4 sequences from Change Detection dataset. Columns (a) and (b) show the original query image and the ground truth (GT) for the foreground. Columns (c) and (f) show the results of COROLA for estimating the background $L$, and the detected foreground objects $\textsc{s}$, respectively. Columns (d) and (e) show intermediate results for outliers $E$, and $\hat{E}$, respectively. }}
\label{fig:qualitative_s2}
\end{figure}

\begin{figure}[t]
\centering
\includegraphics[scale=0.6]{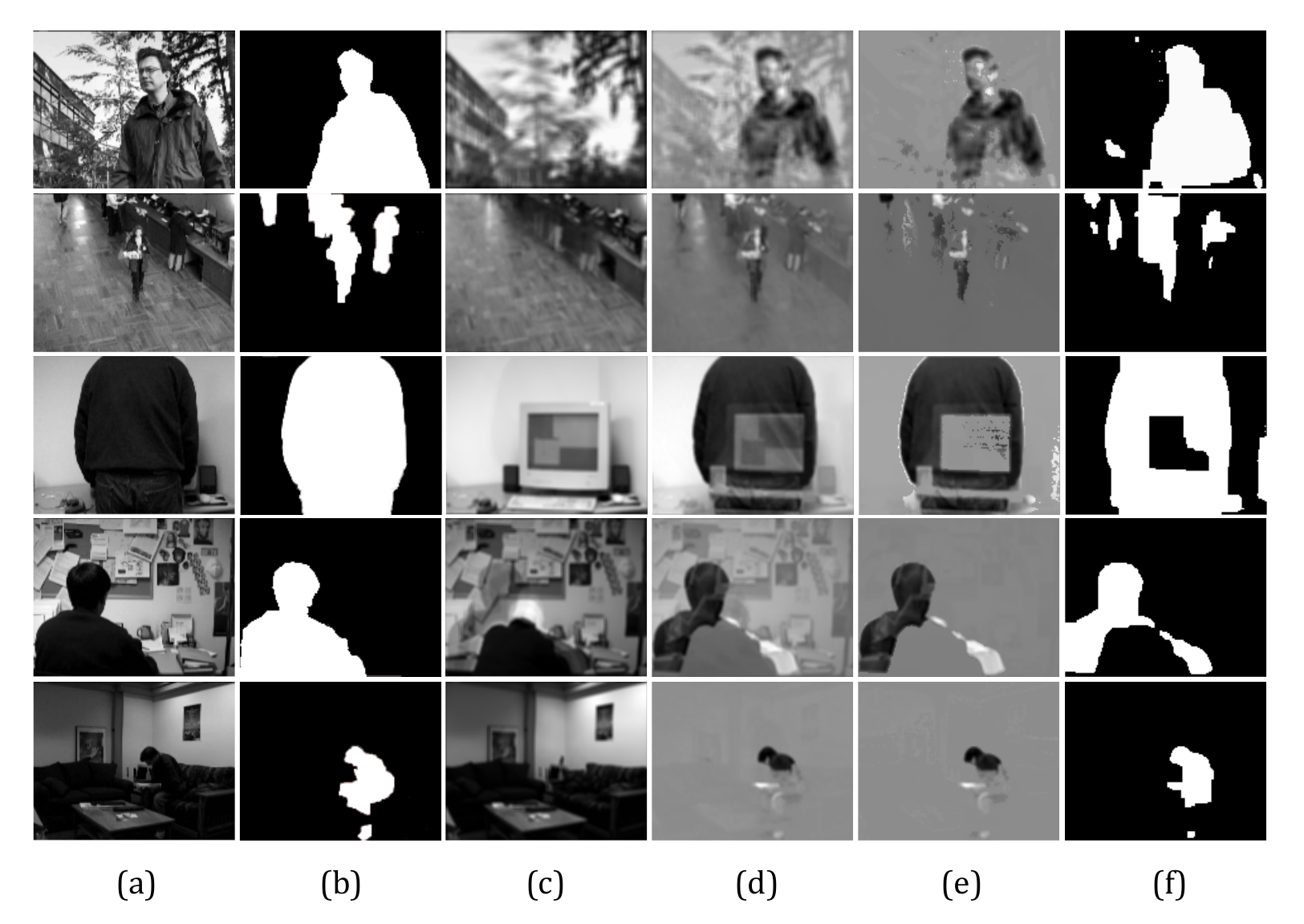}
\caption{\footnotesize{The results of COROLA on 5 sequences from Wallflower dataset. Columns (a) and (b) show the original query image and the ground truth (GT) for the foreground. Columns (c) and (f) show the results of COROLA for estimating the background $L$, and the detected foreground objects $S$, respectively. Columns (d) and (e) show intermediate results for outliers $E$, and $\hat{E}$, respectively. }}
\label{fig:qualitative_s3}
\end{figure}

Figs.~\ref{fig:qualitative_s1},~\ref{fig:qualitative_s2}, and~\ref{fig:qualitative_s3} show the qualitative results of COROLA for background estimation and foreground detection for all sequences of Table~\ref{table_static_dataset} from three datasets I2R, Change Detection, and Wallflower, respectively. Figs.~\ref{fig:qualitative_s1},~\ref{fig:qualitative_s2}, and~\ref{fig:qualitative_s3} also shows the role of GMM to separate outliers from noise. These results are shown in columns (d) and (e) as $E$, and $\hat{E}$,  respectively.
The results in Figs.~\ref{fig:qualitative_s1},~\ref{fig:qualitative_s2}, and~\ref{fig:qualitative_s3} demonstrate the capability of COROLA to detect moving objects and background modelling accurately. The estimated background in the first row of Fig.~\ref{fig:qualitative_s1} has some ghost because the input image is the $23^{rd}$ of the sequence and the parameters have not been learned well enough to build an accurate background. In general, for short sequences the computed background model by a batch method such as DECOLOR is more accurate than COROLA because online methods need sufficient samples for training to be stable. However, for long sequences COROLA can provide comparable results with batch methods.

We also compare COROLA quantitatively with competing online and batch methods. Table~\ref{table_COROLA_other1} compares COROLA with MOG, GRASTA, OPRMF, and ORPCA in terms of F-measure. In most of the cases COROLA works much better than all other online methods, specifically in very noisy and dynamic scenes such as Fountain, Campus, Canoe, Fall, and Fountain02 sequences. Because in these sequences moving parts of background are often classified as foreground in other online methods. In contrast, COROLA is able to deal with the difficult background conditions. By using GMM and (\ref{eq:9}) the difference between outliers and the rest of pixels is boosted and this allows COROLA to detect intermittently moving objects better than other competing online methods.

To show the capability of COROLA, we have also included ``OR-PCA+MRF" \cite{Related_ORPCA_MRF} in our evaluation. Even though this method sets manually all parameters for each sequence, since this approach does not use an optimization framework, it does not perform as well as COROLA in most of the sequences.

Table~\ref{table_COROLA_other2} compares COROLA with IALM, FPCP, GoDec, SSGODec, APG, and DECOLOR, which are fast and accurate batch methods in the literature, in terms of F-measure. For some sequences such as Fountain, Campus, Canoe, Fountain02, Overpass, and TimeOfDay COROLA works much better than other methods. Because in some of these sequences, background is very noisy (i.e. Campus and Fountain02), the constraints of connectedness and sparseness on the subspace of images prove to be useful, which both DECOLOR and COROLA methods exploit leading to much better results than other methods. Further, in some cases the objects move very slowly (i.e. Canoe) or stop for a long time (Overpass, Fountain, and TimeOfDay) none of the competing methods can produce accurate results. In contrast, COROLA produces acceptable results for these challenging sequences for the same reasons as for the results of Table~\ref{table_COROLA_other1}, i.e., using GMM and (\ref{eq:9}) the difference between outliers and the rest of pixels is boosted and so COROLA can detect intermittently moving objects better than other methods.
In summary, Tables~\ref{table_COROLA_other1} and~\ref{table_COROLA_other2} convincingly demonstrate that our method outperforms the state-of-the-art in terms of F-measure.

\setlength{\tabcolsep}{2pt}
\begin{table}[!t]
\scriptsize
\begin{center}
\caption{\footnotesize{Comparison of F-measure score between COROLA and online methods}}
\label{table_COROLA_other1}
\begin{tabular}{lllllllllllll}
\hline\noalign{\smallskip}
Sequence          && MOG    && GRASTA && OPRMF  && ORPCA  && ORPCA &&COROLA\\
                          &&              &&                &&                &&                &&   +MRF &&                \\
\cline{1-13}
WaterSurface      && 0.4723 && 0.7531 && 0.5483 && 0.6426 &&  0.9166 && \textbf{0.9503}\\
Fountain          && 0.7766 && 0.4978 && 0.2393 && 0.2870 &&  0.8283 && \textbf{0.9175}\\
Curtain           && 0.7709 && 0.7046 && 0.4199 && 0.8504 &&  0.8920 && \textbf{0.9038}\\
Hall              && 0.5802 && 0.7471 && 0.7215 && 0.7329 &&  0.7844 && \textbf{0.8298}\\
Campus            && 0.4510 && 0.1885 && 0.1700 && 0.1893 &&   --    && \textbf{0.7650}\\
Escalator         && 0.3869 && 0.5474 && 0.5179 && 0.4452 &&   --    && \textbf{0.7714}\\
Lobby             && 0.5628 && 0.8231 && 0.6728 && 0.6336 &&  0.8081 && \textbf{0.8129}\\
ShoppingMall      && 0.5275 && 0.6816 && 0.6621 && 0.5541 &&   --    && \textbf{0.7452}\\
Canoe             && 0.5114 && 0.5386 && 0.4400 && 0.5152 &&  0.8534 && \textbf{0.8901}\\
Fall              && 0.5420 && 0.5057 && 0.4929 && 0.4030 &&   --    && \textbf{0.8596}\\
Fountain02        && 0.7801 && 0.3569 && 0.2926 && 0.4684 &&  0.8517 && \textbf{0.8642}\\
Overpass          && 0.5095 && 0.5609 && 0.5105 && 0.6079 &&  0.8272 && \textbf{0.8471}\\
WavingTrees       && 0.6639 && 0.7354 && 0.5259 && 0.6315 &&  \textbf{0.8689} && 0.8688\\
Bootstrap         && 0.4613 && 0.5635 && 0.5627 && 0.5619 &&   --    && \textbf{0.6930}\\
Camouflage        && 0.6922 && 0.2191 && 0.6525 && 0.2307 &&  \textbf{0.9118} && 0.8738\\
ForegroundAperture&& 0.2601 && 0.6757 && 0.5628 && 0.6118 &&  \textbf{0.6910} && 0.6709\\
%LightSwitch       &&        && 0.5291 && 0.4393 && 0.7757 && 0.3038\\
TimeOfDay         && 0.6147 && 0.5645 && 0.5258 && 0.6315 &&   --    && \textbf{0.8344}\\

%Boring parking && 0.6945 && 7865 && 70.20 && 0.8916 && 0.9236\\
%Beware of the trains && 0.7793 &&
\cline{1-13}
\hline
\end{tabular}
\end{center}
\end{table}
\setlength{\tabcolsep}{1.4pt}

\setlength{\tabcolsep}{2pt}
\begin{table}[!t]
\scriptsize
\begin{center}
\caption{\footnotesize{Comparison of F-measure score between COROLA and batch methods}}
\label{table_COROLA_other2}
\begin{tabular}{lllllllllllllll}
\hline\noalign{\smallskip}
Sequence          && IALM   &&  FPCP  && GoDec  && SSGoDec&&  APG   && DECOLOR && COROLA\\
\cline{1-15}
WaterSurface      && 0.3519 && 0.4910 && 0.4304 && 0.4473 && 0.5907 && 0.9022  && \textbf{0.9503}\\
Fountain          && 0.1633 && 0.1894 && 0.1531 && 0.2574 && 0.2641 && 0.2075  && \textbf{0.9175}\\
Curtain           && 0.3184 && 0.5290 && 0.3706 && 0.4344 && 0.7260 && 0.8700  && \textbf{0.9038}\\
Hall              && 0.5716 && 0.7295 && 0.7128 && 0.5713 && 0.7601 && 0.8169  && \textbf{0.8298}\\
Campus            && 0.1660 && 0.1701 && 0.1640 && 0.1649 && 0.1979 && \textbf{0.7811}  && 0.7650\\
Escalator         && 0.5066 && 0.5192 && 0.1316 && 0.5075 && 0.5440 && \textbf{0.8205}  && 0.7714\\
Lobby             && 0.3213 && 0.7188 && 0.7393 && 0.6194 && 0.7286 && 0.6579  && \textbf{0.8129}\\
ShoppingMall      && 0.6093 && 0.6256 && 0.6143 && 0.5880 && 0.7057 && 0.6382  && \textbf{0.7452}\\
Canoe             && 0.5072 && 0.5169 && 0.5107 && 0.3091 && 0.4193 && 0.1603  && \textbf{0.8901}\\
Fall              && 0.4112 && 0.4191 && 0.4137 && 0.4236 && 0.5232 && \textbf{0.8760}  && 0.8596\\
Fountain02        && 0.2553 && 0.3066 && 0.2713 && 0.2714 && 0.3204 && 0.8327  && \textbf{0.8642}\\
Overpass          && 0.5492 && 0.5528 && 0.5454 && 0.5517 && 0.5698 && 0.3573  && \textbf{0.8471}\\
WavingTrees       && 0.5130 && 0.5130 && 0.5113 && 0.1829 && 0.7031 && \textbf{0.8845}  && 0.8688\\
Bootstrap         && 0.6517 && 0.6525 && 0.6490 && 0.5567 && 0.5619 && 0.6342  && \textbf{0.6930}\\
Camouflage        && 0.6518 && 0.6518 && 0.6428 && 0.6426 && 0.3441 && 0.3661  && \textbf{0.8738}\\
F-A               && 0.3233 && 0.3238 && 0.3238 && 0.6854 && \textbf{0.7200} &&   --    && 0.6709\\
%LightSwitch       && 0.3471 && 0.3653 && 0.1817 && 0.7973 && 0.8347 &&         && 0.3038\\
TimeOfDay         && 0.1523 && 0.2187 && 0.1630 && 0.1664 && 0.6808 && 0.4683  && \textbf{0.8344}\\

%Boring parking && 0.6945 && 7865 && 70.20 && 0.8916 && 0.9236\\
%Beware of the trains && 0.7793 &&
\cline{1-11}
\hline
\end{tabular}
\end{center}
\end{table}
\setlength{\tabcolsep}{1.4pt}

\subsubsection{Computational Cost}
COROLA is implemented in Matlab and C++. We run all experiments on a PC with a 3.4 GHz Intel i7 CPU and 16 GB RAM. To show the importance of online methods in continuous operation we compare the scalability of COROLA with DECOLOR under varying spatial resolution and the number of images.
%Fig.~\ref{fig:Computation_time} shows the average computation time of DECOLOR with different number of images.

%\begin{figure}[b]
%\centering
%\includegraphics[scale=0.35]{Computationtime.eps}
%\caption{Computation time of DECOLOR model with different number of images (x-axis).}
%\label{fig:Computation_time}
%\end{figure}

Unlike DECOLOR, the computational cost of COROLA is independent of the number of images because the dominant cost of DECOLOR comes from the computation of SVD in each iteration. By increasing the size of the matrix $D$, the computation time of DECOLOR grows at least linearly with respect to the number of images. We compare the computation time of COROLA with DECOLOR after convergence of both methods in Table~\ref{compare_time}. In this table, the average time for processing of each frame by DECOLOR increases where it is an order of magnitude slower than COROLA for sequences longer than 1000 images.

\setlength{\tabcolsep}{3pt}
\begin{table}[t]
\begin{center}
\caption{Time evaluation of COROLA with DECOLOR method}
\label{compare_time}
\begin{tabular}{lllllllll}
\hline\noalign{\smallskip}
Methods && Resolution $\times$ \#images && Low Rank (s) && MRF (s) && Total (s)\\
\cline{1-9}
        && $[320 \times 240] \times 200$ && 0.1036 && 0.0828 && 0.1864\\
        && $[320 \times 240] \times 400$ && 0.1531 && 0.1297 && 0.2828\\
DECOLOR && $[320 \times 240] \times 600$ && 0.1687 && 0.1601 && 0.3279\\
        && $[320 \times 240] \times 800$ && 0.2016 && 0.1825 && 0.3841\\
        && $[320 \times 240] \times 1000$ && 0.3948 && 0.3191 && 0.7139\\
\hline
COROLA    && $[320 \times 240] \times 1000$ && 0.0231 && 0.0605 && 0.0836\\
\cline{1-9}
\hline
\end{tabular}
\end{center}
\end{table}
\setlength{\tabcolsep}{1.4pt}

Scalability in spatial resolution is another advantage of online method against batch processing methods. Increasing the resolution of images significantly affects DECOLOR method. Using high resolution images results in a huge matrix $D$ so that decomposing $D$ becomes very expensive. On the other hand, COROLA is an online method and is independent from the number of images, i.e., we do not have to deal with a large $D$ and its computation time grows only with the image resolution.

%\begin{figure}[b]
%\begin{center}$
%\begin{array}{ccc}
%\includegraphics[scale=0.3]{Computation_time_cars6.pdf}&
%\includegraphics[scale=0.3]{Computation_time_cars7.pdf}&
%\includegraphics[scale=0.3]{Computation_time_people1.pdf}
%\end{array}$
%\end{center}
%\caption{Comparison of computation time in different resolution}
%\label{time_comparing}
%\end{figure}

\subsection{Experiments on a Moving Camera}

In this section, we test our method on real public sequences for moving cameras namely ``Berkeley motion segmentation dataset"~\cite{dataset_motion}. Table~\ref{moving_camera} shows the details of five challenging sequences that we use in our experiments.
\setlength{\tabcolsep}{3pt}
\begin{table}[b]
\begin{center}
\caption{Details of all sequences used in our experiments for moving camera}
\label{moving_camera}
\begin{tabular}{lllllll}
\hline\noalign{\smallskip}
Dataset &&& Sequences &&& Size $\times$ \#frames\\
\cline{1-7}
&&& cars6     &&& [320,240] $\times$ 30\\
&&&cars7      &&& [320,240] $\times$ 24\\
Berkeley motion segmentation &&& people1 &&& [320,240] $\times$ 40\\
&&& tennis   &&& [320,240] $\times$ 200\\
&&& marple13 &&& [320,240] $\times$ 75\\
\hline
\end{tabular}
\end{center}
\end{table}
\setlength{\tabcolsep}{1.4pt}

\begin{figure}[t]
\centering
\includegraphics[scale=0.7]{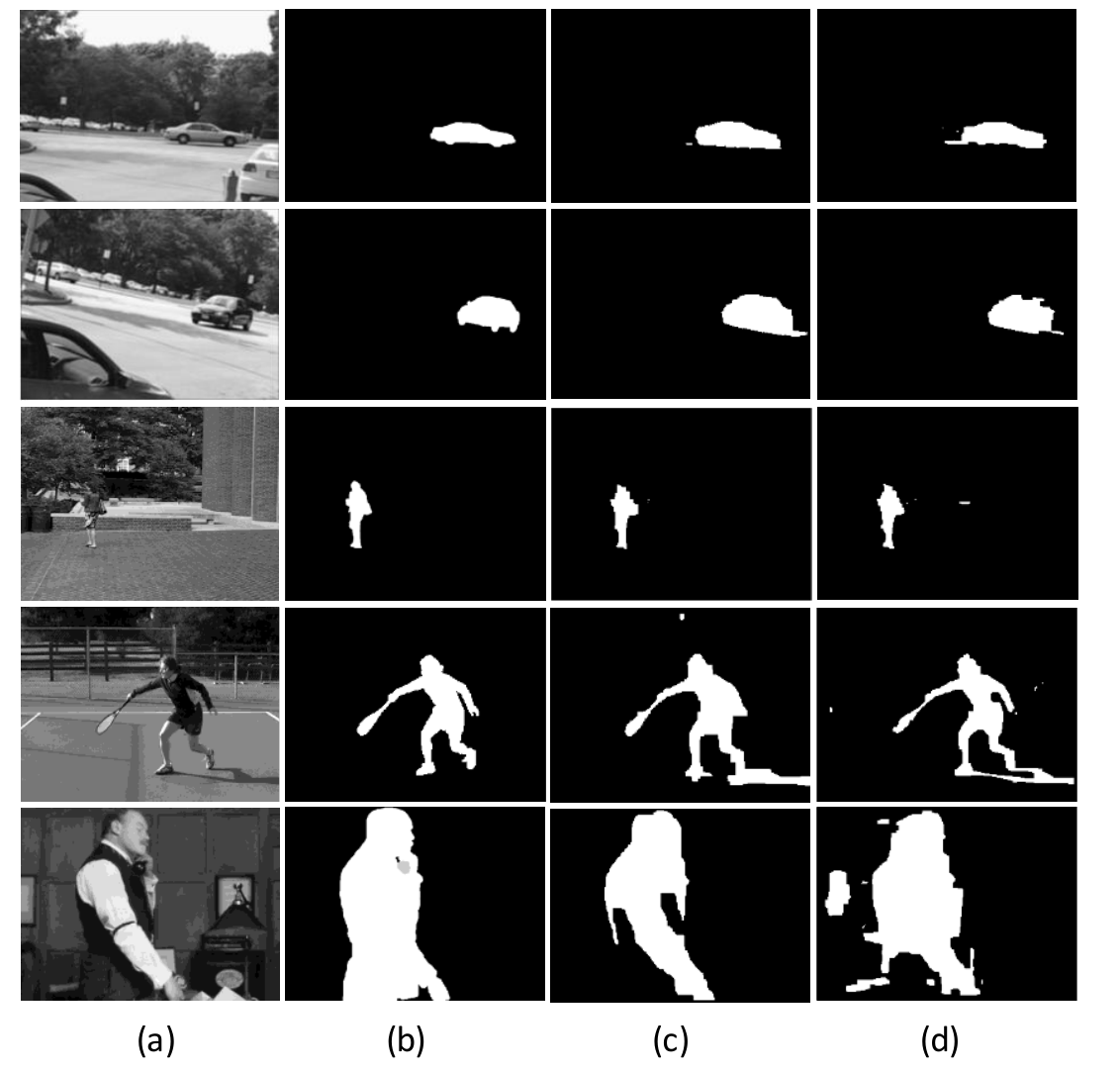}
\caption{\footnotesize{Comparison of foreground objects between DECOLOR and COROLA. columns (a) and (b) show the input image and its ground truth. columns (c) and (d) show the obtained foreground mask for DECOLOR and COROLA methods, respectively.  }}
\label{fig:qualitative_m1}
\end{figure}
We compare our method with DECOLOR as the leading method based on low-rank approximation that can handle the problem of object detection with a moving camera in a short sequence. Although recently, He {\it{et al.}}~\cite{t-GRASTA} has proposed transformed-GRASTA, it only works well for camera jitter and it is not appropriate for moving camera.
Fig.~\ref{fig:qualitative_m1} shows the qualitative results of COROLA in comparison with DECOLOR method for moving object detection using a moving camera. First three experiments have been performed on short sequences ``cars6", ``cars7", ``people1" and the results from COROLA are comparable with those from DECOLOR method. For the last two sequences ``marple13" and ``Tennis", DECOLOR has a problem to align images when the last images are not similar with the first images of these sequences. This is common in continuous processing and all of batch methods have problem with this. To show the result of DECOLOR on marple13 and tennis sequences (in the last two rows of Fig.~\ref{fig:qualitative_m1}), we used last 30 images of the sequences, which have less camera motion. Since the last images in the sequence are no longer similar to the initial ones in the matrix, DECOLOR failed, as expected. In contrast, since COROLA works online and only considers the last two images it can process the last two sequences of Table~\ref{moving_camera} without any problems and provides acceptable results in comparison with DECOLOR. For completeness, we have also included in our comparative study another online registration based method in~\cite{int5}.

Table~\ref{moving_quantitative} shows the quantitative evaluation of COROLA in comparison with DECOLOR and the method in~\cite{int5}. Experiments over all five sequences show that the results of COROLA is comparable with DECOLOR for the last 30 images of a sequence but has the advantage in terms of its ability for real-time continuous processing. With more than 30 images in a sequence, DECOLOR can no longer produce a valid result due to the significant dissimilarity of the images later in the sequence from the initial ones.  In contrast, our sequential method is always able to produce a valid result often with higher accuracy.

\setlength{\tabcolsep}{3pt}
%\begin{savenotes}
\begin{table}[t]
\begin{center}
\caption{Comparison of F-measure score. $^1$ last 30 images is used}
\label{moving_quantitative}
\begin{tabular}{lllllllllllll}
\hline\noalign{\smallskip}
Sequence &&&     &&& FFD based model &&& DECOLOR &&& COROLA\\
\cline{1-13}
cars6    &&&     &&& 0.8870 &&& 0.9052  &&& \textbf{0.9409}\\
cars7    &&&     &&& 0.8257 &&& 0.8441  &&& \textbf{0.8867}\\
people1  &&&     &&& 0.8122 &&& \textbf{0.9666}  &&& 0.9056\\
tennis   &&&     &&& 0.8494 &&& $0.8404^1$/NA &&& \textbf{0.8642}\\
marple13 &&&     &&& 0.6407 &&& $0.8063^1$/NA  &&& \textbf{0.8271}\\
\cline{1-11}
\hline
\end{tabular}
\end{center}
\end{table}
%\end{savenotes}
\setlength{\tabcolsep}{1.4pt}

\section{Conclusion}
\label{section5}
In this paper, we have proposed a novel online method named COROLA to detect moving objects in a video using the framework of low-rank matrix approximation. Our online framework works iteratively on each image of the video to extract foreground objects accurately. The key to our online formulation is to exploit the sequential nature of a continuous video of a scene where the background model does not change discontinuously and can therefore be obtained by updating the background model learned from preceding images. We have applied COROLA to the case of a moving camera. Since our method works online and is independent of the number of images, it is suitable for real-time object detection in continuous monitoring tasks. Our method overcomes the problems of batch methods in terms of memory storage, time complexity, and camera motion.
Also important to the success of COROLA is using Gaussian model to separate noise from outliers and also to tune the costs of assigning labels in MRF via $\sigma$ and weights of Gaussian parameters, dynamically and automatically especially when the object moves very slow or stops for some frames. Based on our extensive experiments on synthetic data and real data sequences, we are able to establish that COROLA archives the best performance in comparison with all evaluated methods including the state-of-the-art batch and online methods.

Despite its satisfactory performance in all of our experiments, COROLA shares one disadvantage with DECOLOR. Since both methods have non-convex formulations, they might converge to a local minimum with results depending on initialization of parameters; however, for the case of background modeling, images are roughly similar and parameters do not change significantly. Therefore, the issue of local minimum has not affected successful object detection in our experiments. A challenge facing COROLA is severe illumination changes and this is a problem of all online methods. In the future, we plan to develop a version of COROLA that can work under severe illumination changes.

\textbf{Acknowledgements}
This work was supported in part by the Natural Sciences and Engineering Research Counsil
(NSERC) through the NSERC Canadian Field Robotics Network (NCFRN) and by
Alberta Innovates Technology Future (AITF).
%\section{References}
\bibliography{egbib}
\end{document}